\documentclass[10pt,twocolumn,letterpaper]{article}
\usepackage{cvpr}
\usepackage{times}
\usepackage{epsfig}
\usepackage{graphicx}
\usepackage{amsmath}
\usepackage{amssymb}
\usepackage{url}
\usepackage[usenames,dvipsnames]{xcolor}

\usepackage[pagebackref=true,breaklinks=true,letterpaper=true,colorlinks,bookmarks=false]{hyperref}

\renewcommand{\etal}{\textit{et~al.~}}

\graphicspath{{images/}}
\DeclareGraphicsExtensions{.pdf}

\cvprfinalcopy

\def\cvprPaperID{1221}

\ifcvprfinal\fi
\begin{document}

\title{Video Paragraph Captioning Using Hierarchical Recurrent Neural Networks}

\author{ 
  \vspace{-2ex}
  $\text{Haonan Yu}^1$\thanks{This work was done while the
    authors were at Baidu.}
  \enspace$\text{Jiang Wang}^3$
  \enspace$\text{Zhiheng Huang}^{2*}$
  \enspace$\text{Yi Yang}^3$
  \enspace$\text{Wei Xu}^3$\\\\
$^1$Purdue University \hspace{5ex} $^2$Facebook\\
\hspace{3ex}{\tt\small haonanu@gmail.com} 
\hspace{4ex}{\tt\small zhiheng@fb.com}\\
$^3$Baidu Research - Institute of Deep Learning\\
{\tt\small \{wangjiang03,yangyi05,wei.xu\}@baidu.com}\\
}

\maketitle

\begin{abstract}
  We present an approach that exploits hierarchical Recurrent Neural
  Networks (RNNs) to tackle the video captioning problem, i.e.,
  generating one or multiple sentences to describe a realistic video.
  Our hierarchical framework contains a sentence generator and
  a paragraph generator.
  The sentence generator produces one simple short sentence that
  describes a specific short video interval.
  It exploits both temporal- and spatial-attention mechanisms to
  selectively focus on visual elements during generation.
  The paragraph generator captures the inter-sentence dependency by
  taking as input the sentential embedding produced by the sentence
  generator, combining it with the paragraph history, and outputting
  the new initial state for the sentence generator.
  We evaluate our approach on two large-scale benchmark datasets:
  YouTubeClips and TACoS-MultiLevel.
  The experiments demonstrate that our approach significantly
  outperforms the current state-of-the-art methods with BLEU@4 scores
  0.499 and 0.305 respectively.
\end{abstract}

\vspace{-4ex}
\section{Introduction}
\begin{figure*}[t]
  \centering
  \resizebox{\textwidth}{!}{
    \begin{tabular}{cc}
      \begin{tabular}{c}
        \begin{tabular}{c@{\hspace{1ex}}c@{\hspace{1ex}}c@{\hspace{1ex}}c}
          \includegraphics[width=0.25\textwidth]{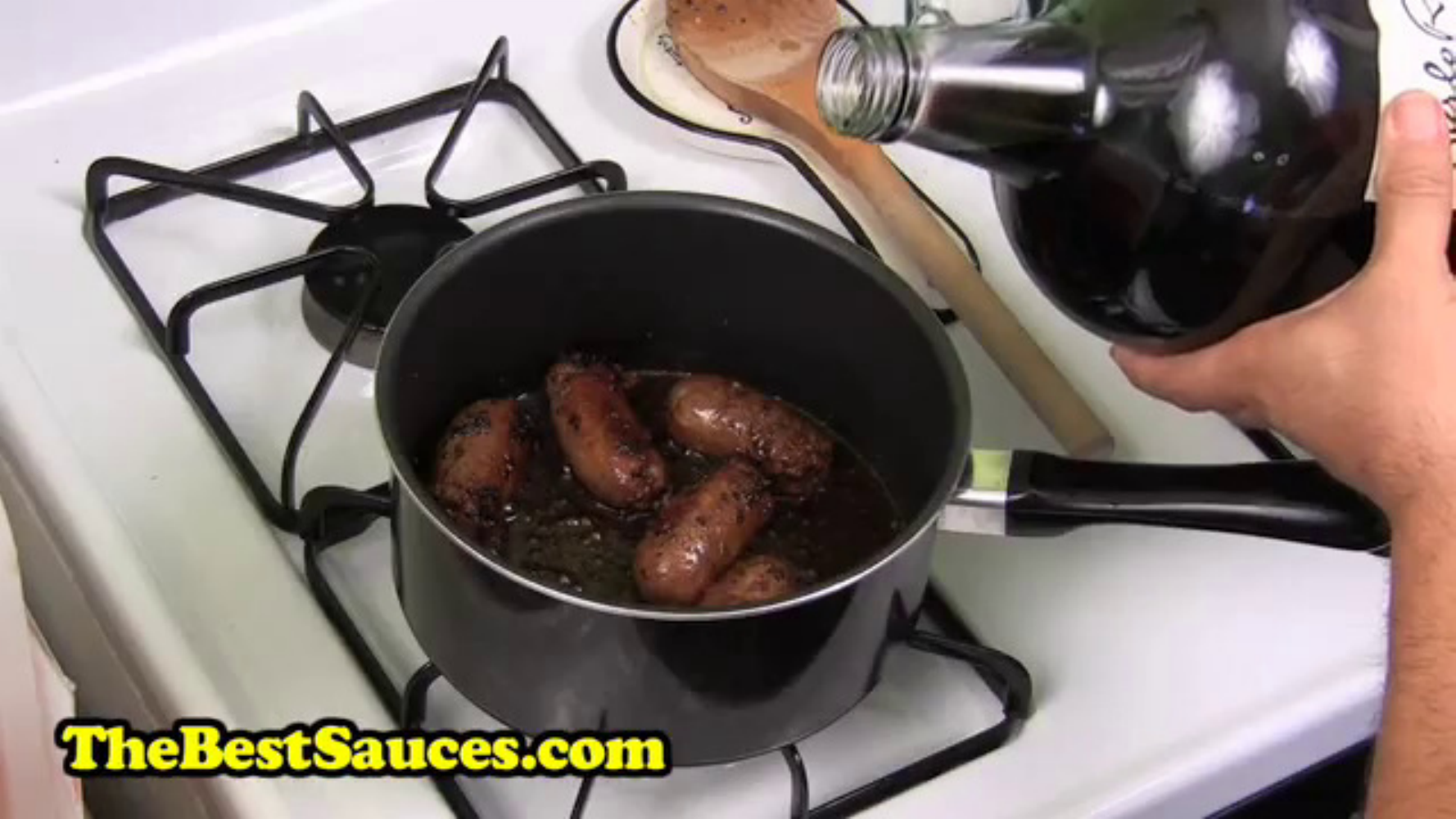}
          &\includegraphics[width=0.25\textwidth]{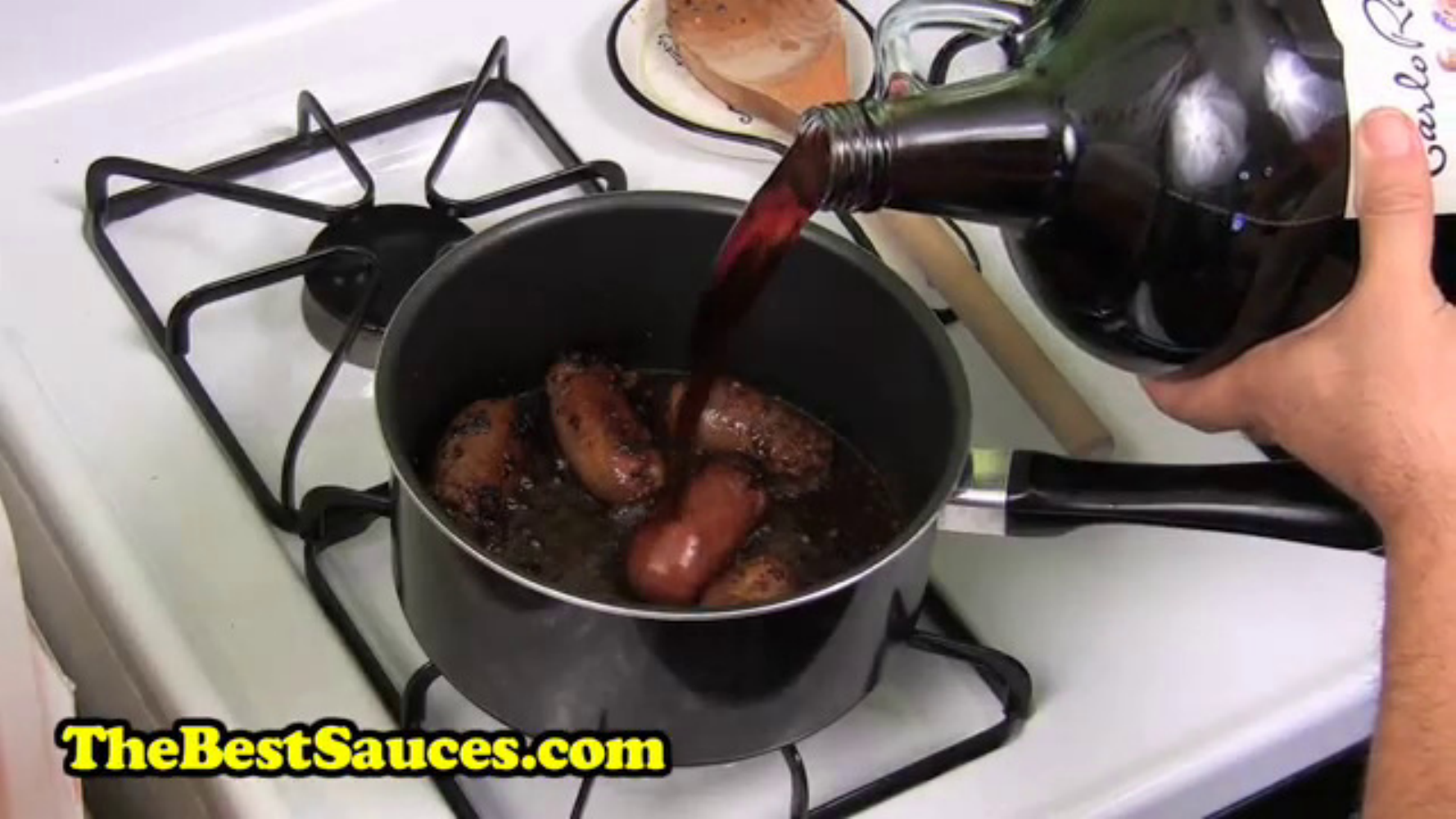}
          &\includegraphics[width=0.25\textwidth]{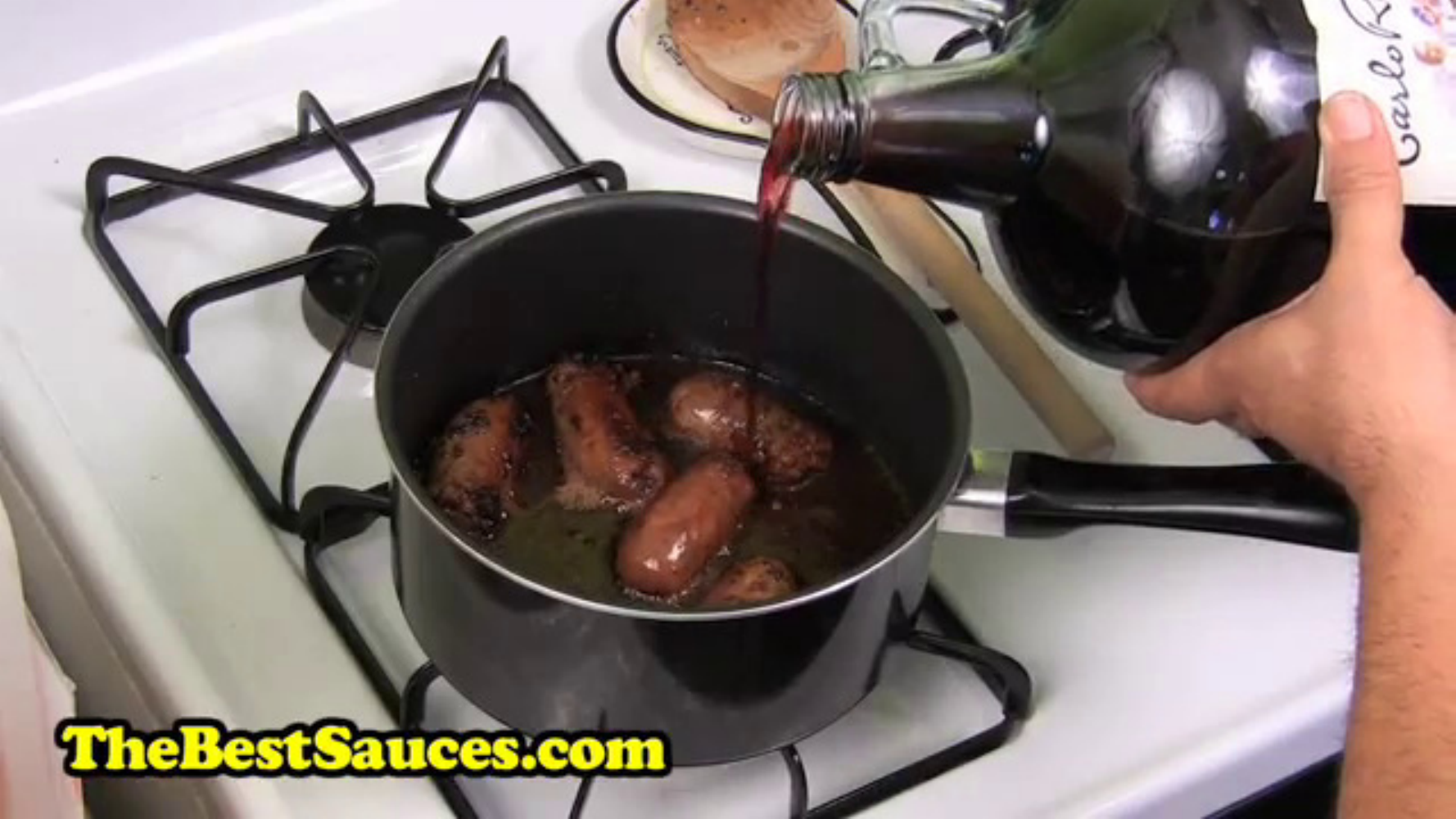}
          &\includegraphics[width=0.25\textwidth]{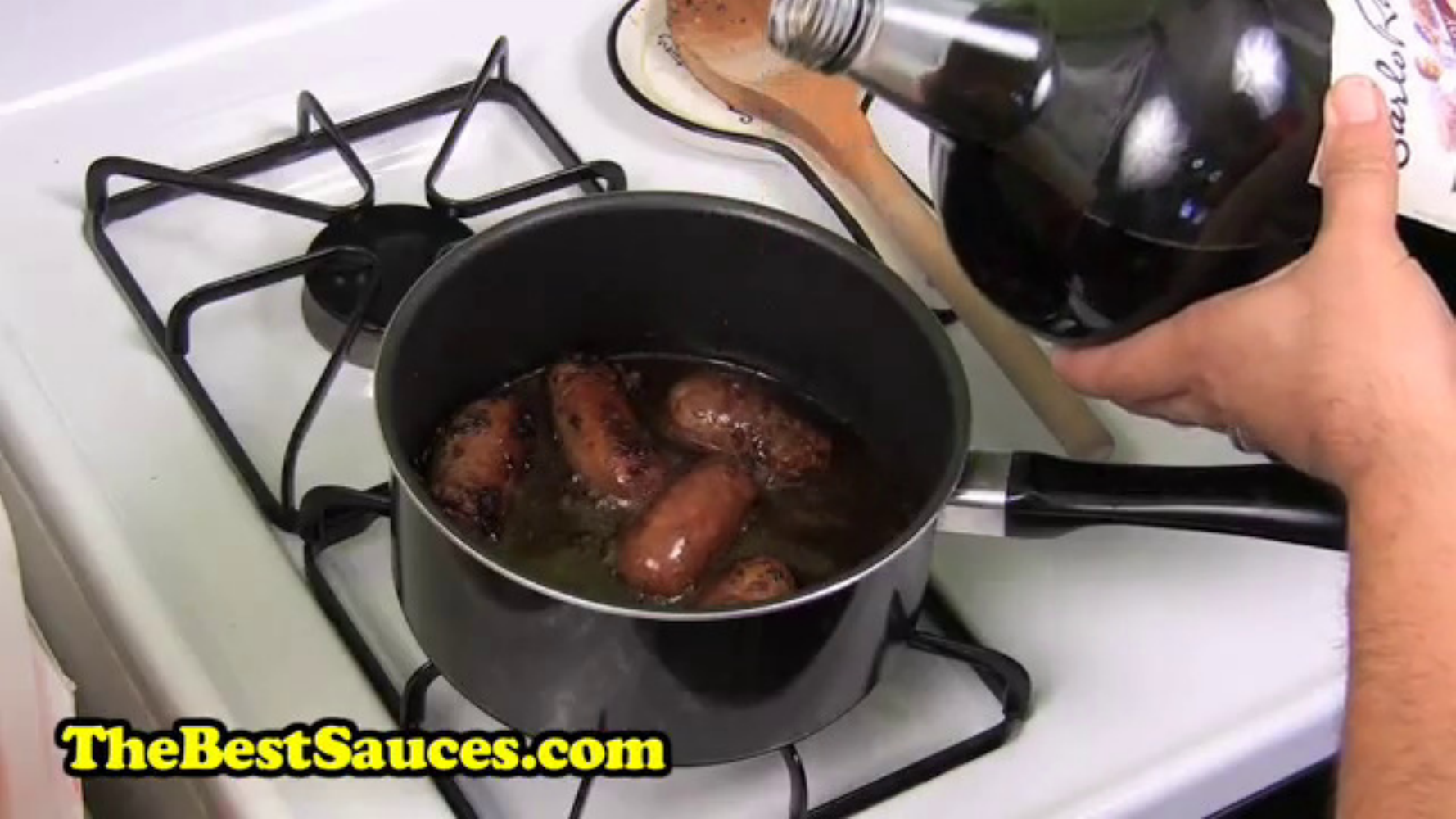}
        \end{tabular}
        \\
        \begin{tabular}{l}
          \textit{\huge{A man is pouring oil into a pot.}}
        \end{tabular}
      \end{tabular}
      &
      \begin{tabular}{c}
        \begin{tabular}{c@{\hspace{1ex}}c@{\hspace{1ex}}c@{\hspace{1ex}}c}
          \includegraphics[width=0.25\textwidth]{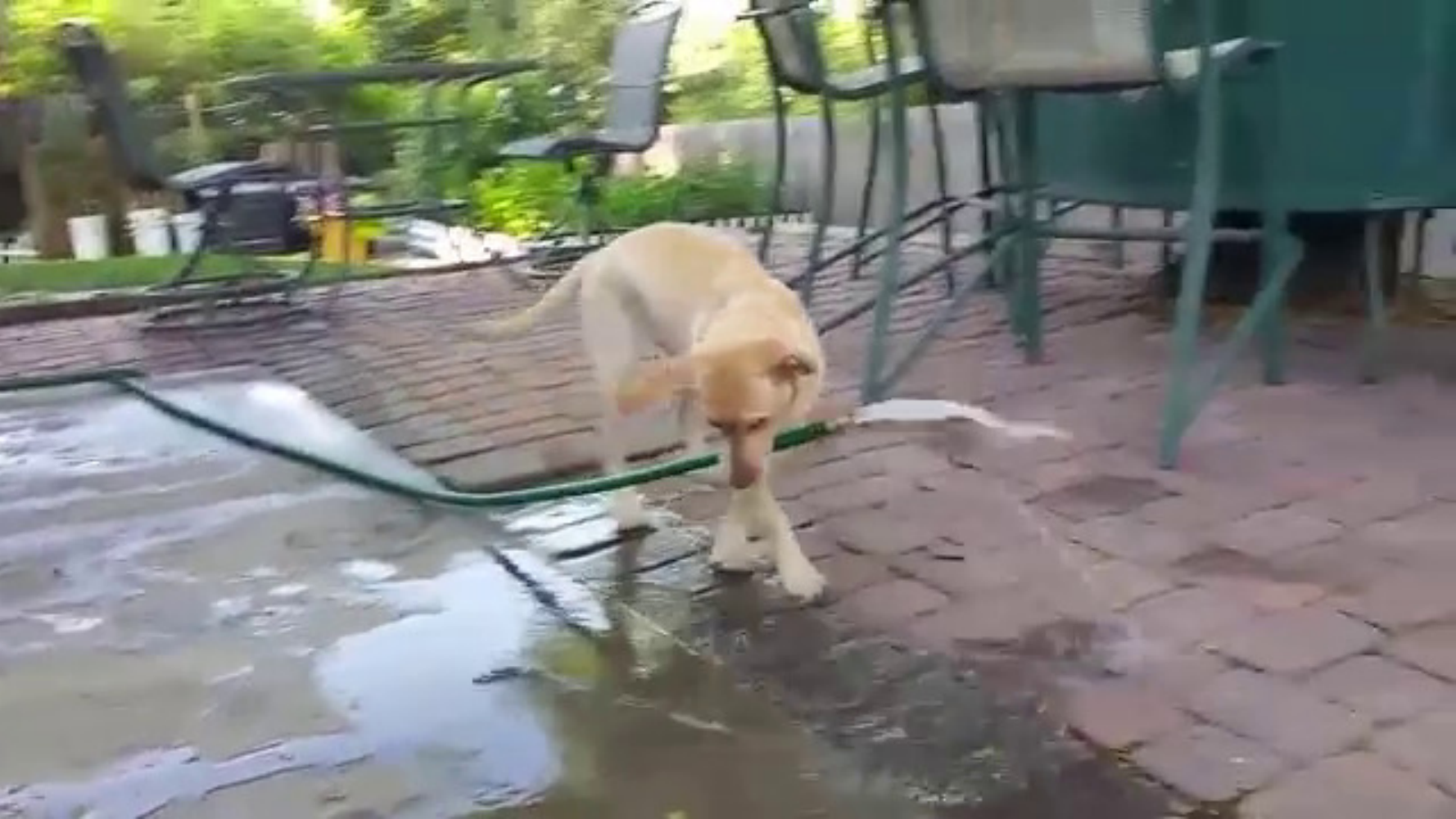}
          &\includegraphics[width=0.25\textwidth]{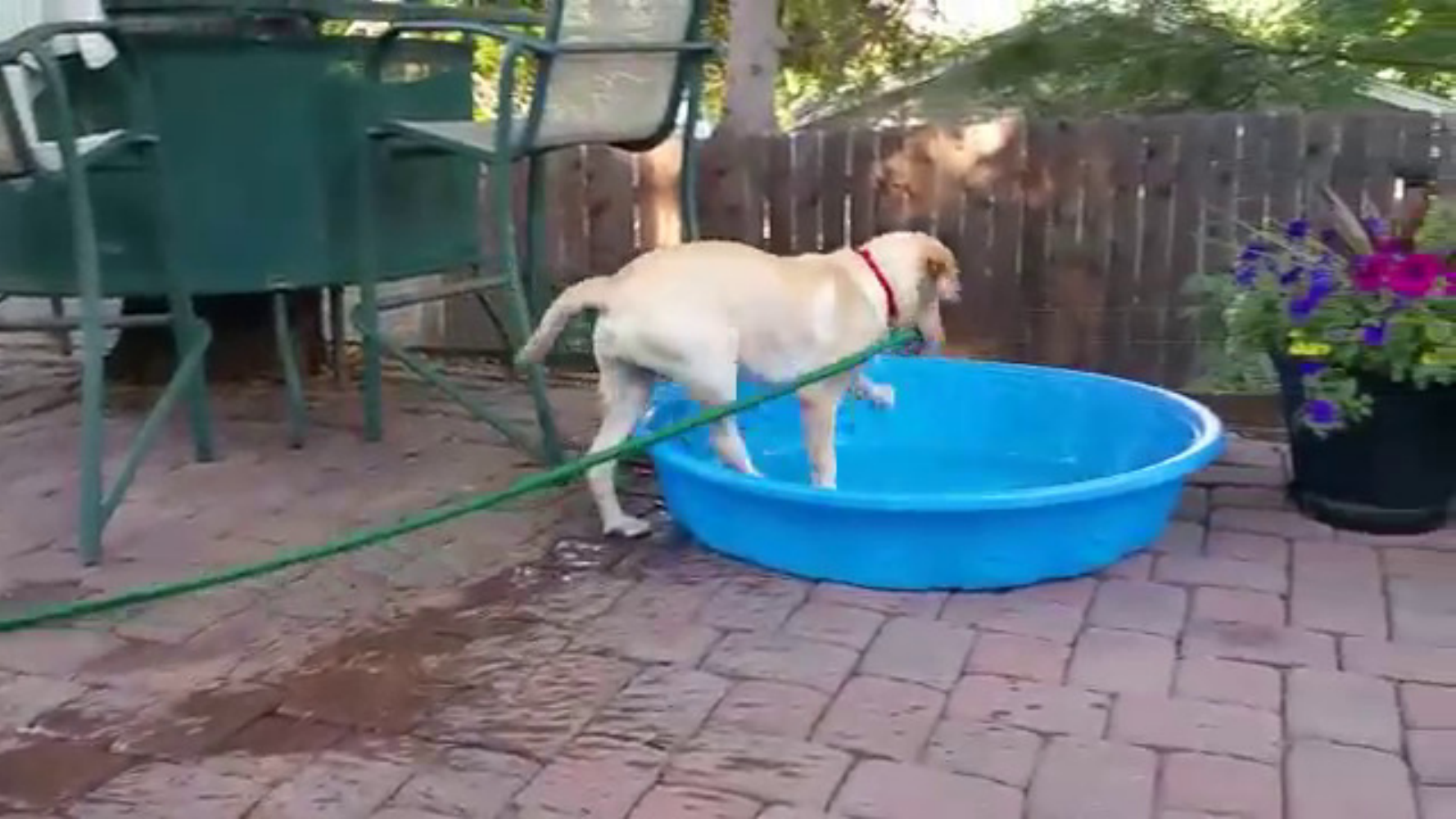}
          &\includegraphics[width=0.25\textwidth]{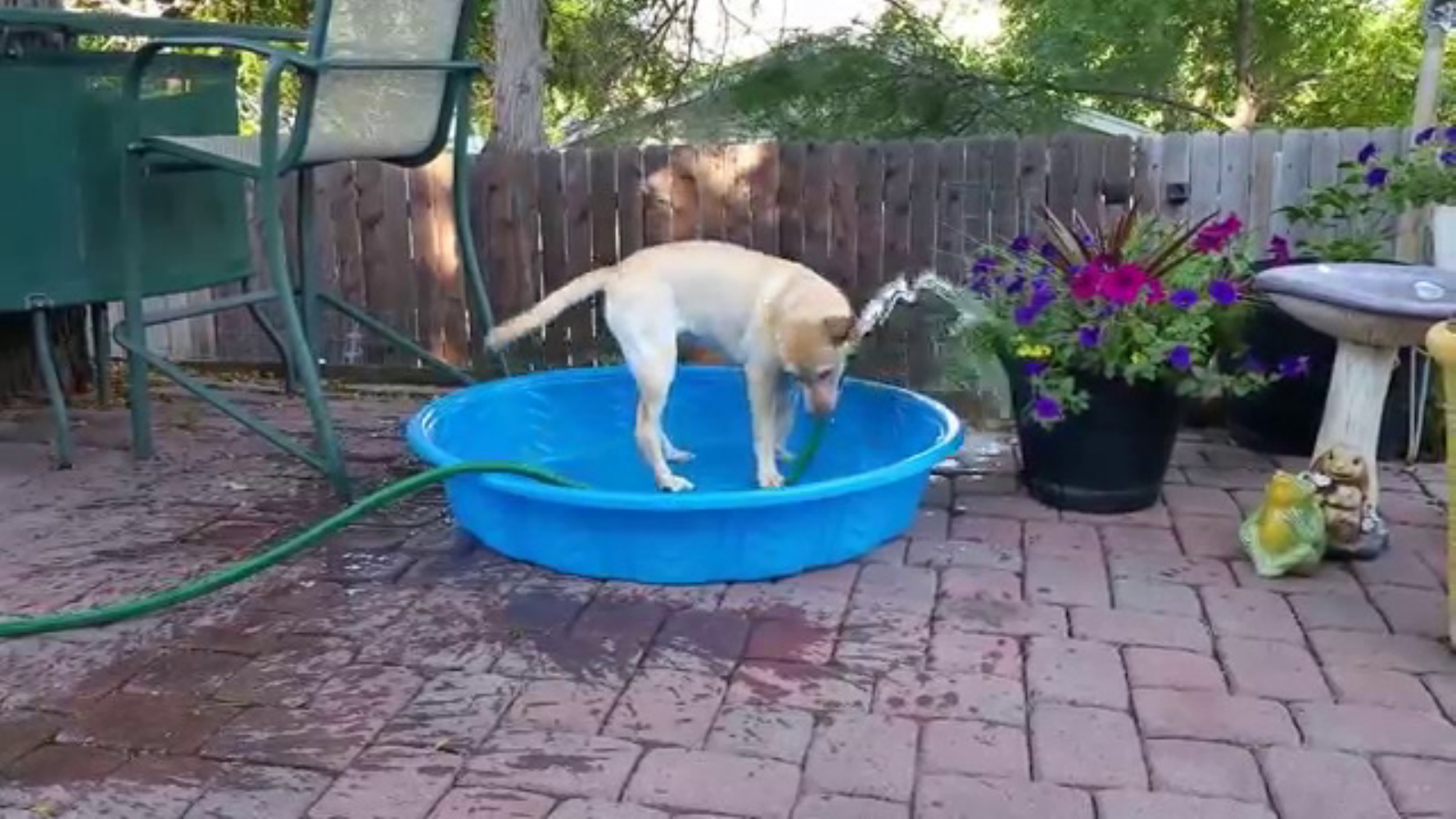}
          &\includegraphics[width=0.25\textwidth]{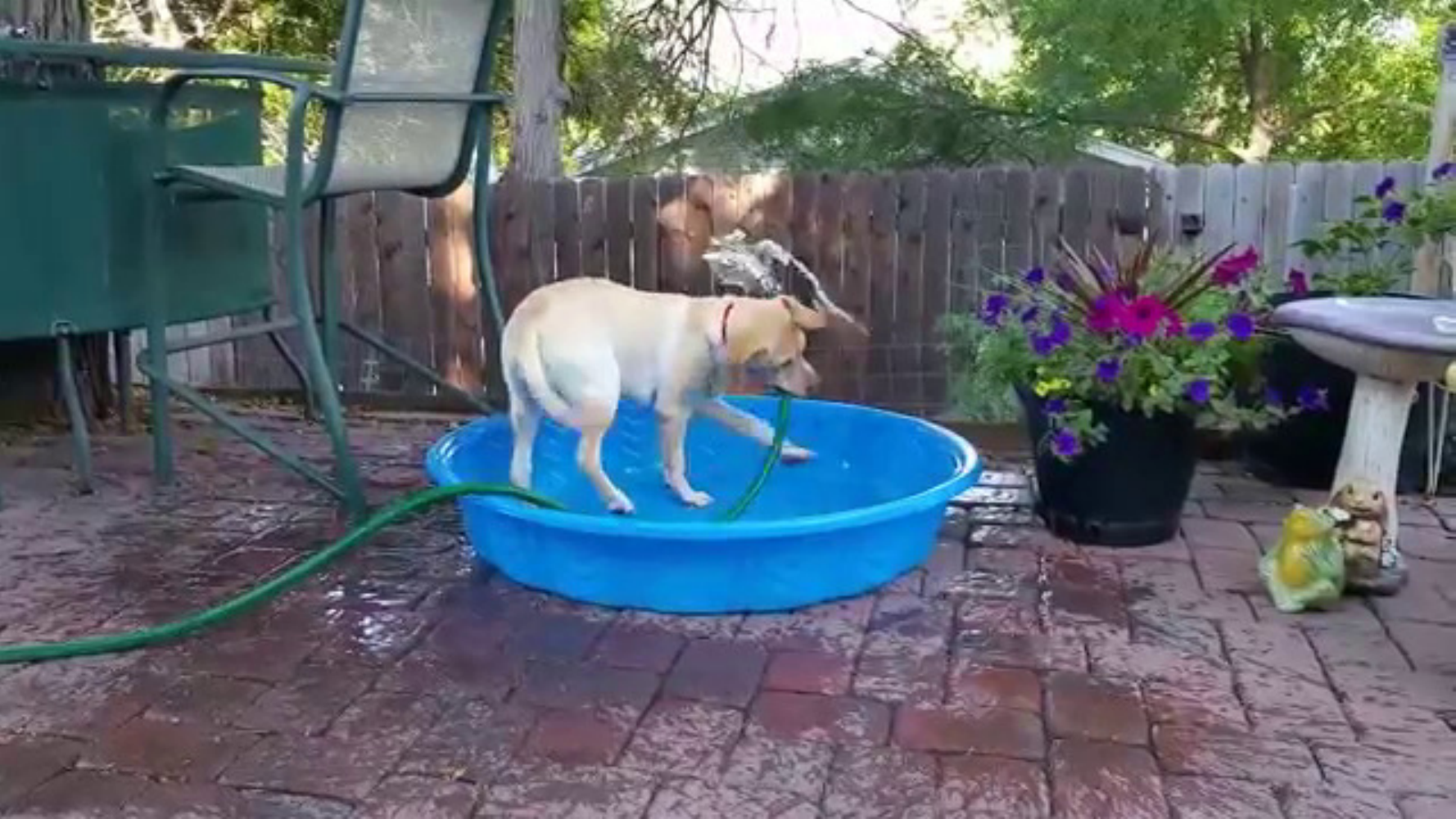}
        \end{tabular}\\
        \begin{tabular}{l}
          \textit{\huge{A dog is playing in a bowl.}}
        \end{tabular}
      \end{tabular}
      \\\\     
      \begin{tabular}{c}
        \begin{tabular}{c@{\hspace{1ex}}c@{\hspace{1ex}}c@{\hspace{1ex}}c@{\hspace{1ex}}c}
          \includegraphics[width=0.2\textwidth]{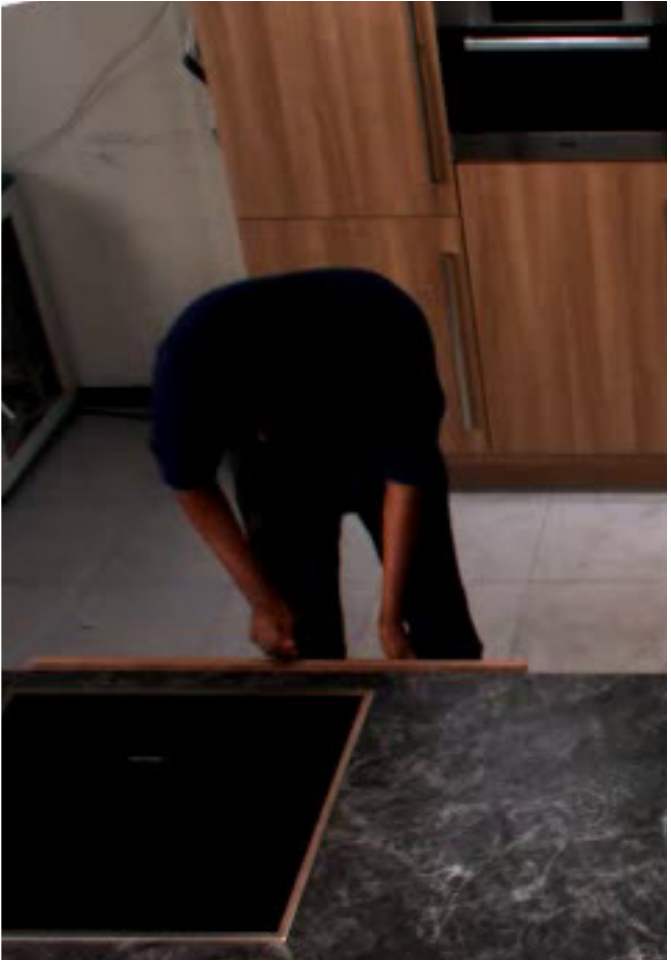}
          &\includegraphics[width=0.2\textwidth]{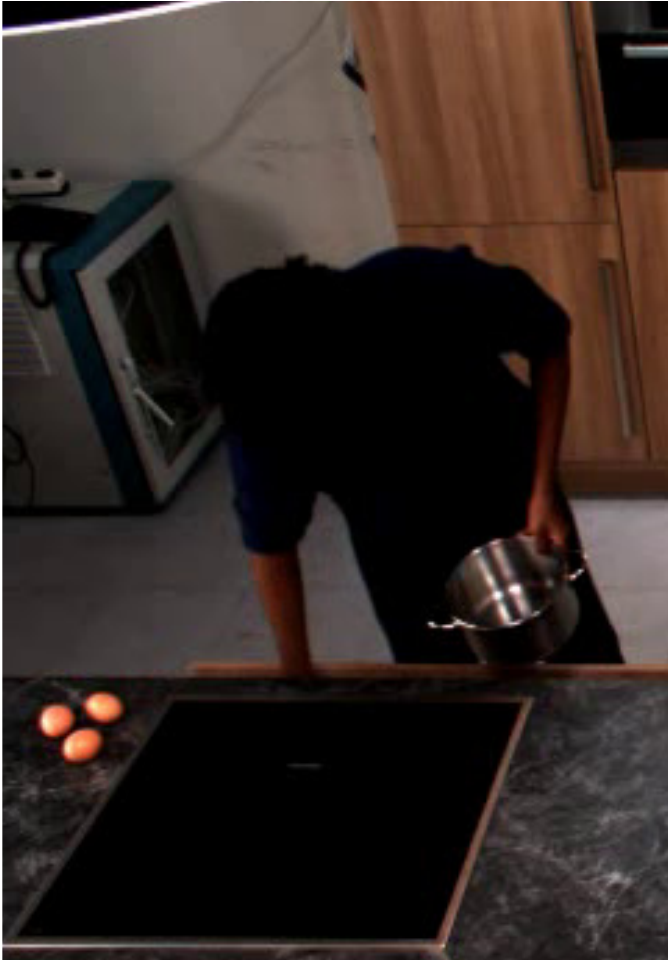}
          &\includegraphics[width=0.2\textwidth]{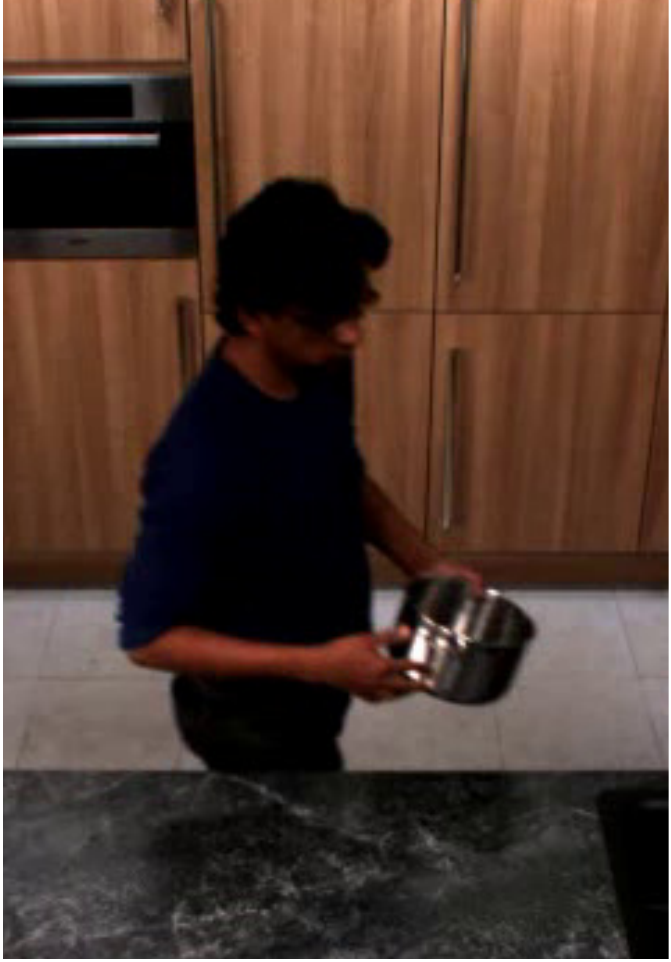}
          &\includegraphics[width=0.2\textwidth]{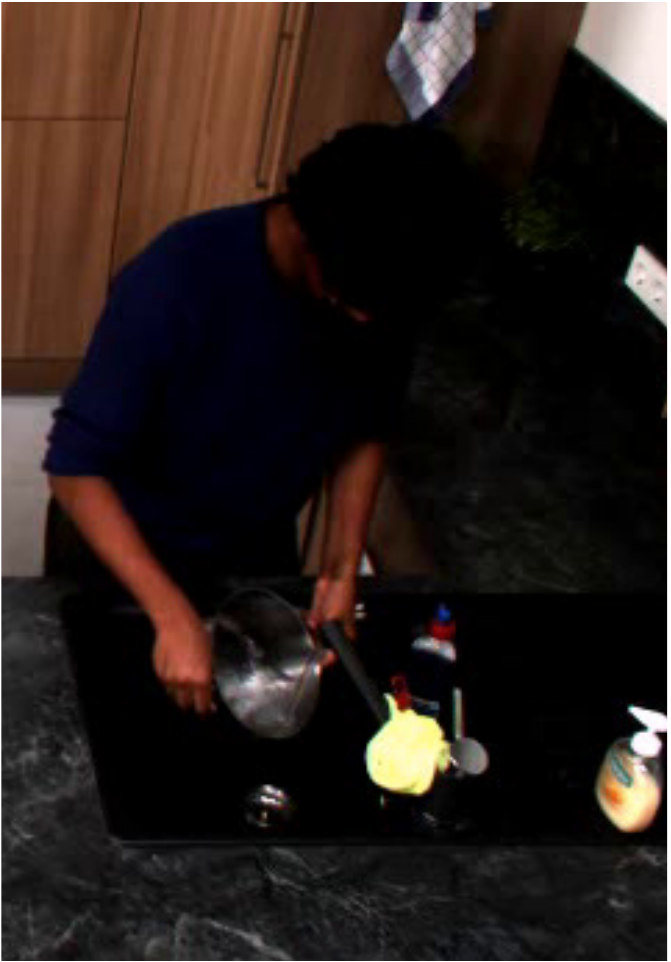}
          &\includegraphics[width=0.2\textwidth]{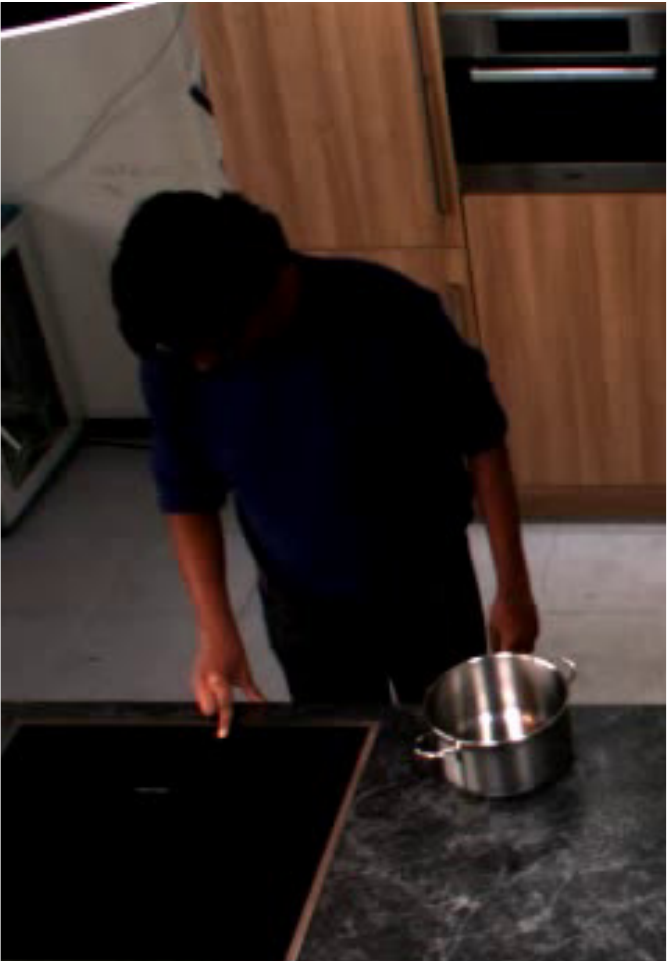}
        \end{tabular}\\
        \begin{tabular}{l}
          \textit{\huge{The person opened the drawer.}}\\
          \textit{\huge{The person took out a pot.}}\\
          \textit{\huge{The person went to the sink.}}\\
          \textit{\huge{The person washed the pot.}}\\
          \textit{\huge{The person turned on the stove.}}\\
        \end{tabular}
      \end{tabular}
      &
      \begin{tabular}{c}
        \begin{tabular}{c@{\hspace{1ex}}c@{\hspace{1ex}}c@{\hspace{1ex}}c@{\hspace{1ex}}c}
          \includegraphics[width=0.2\textwidth]{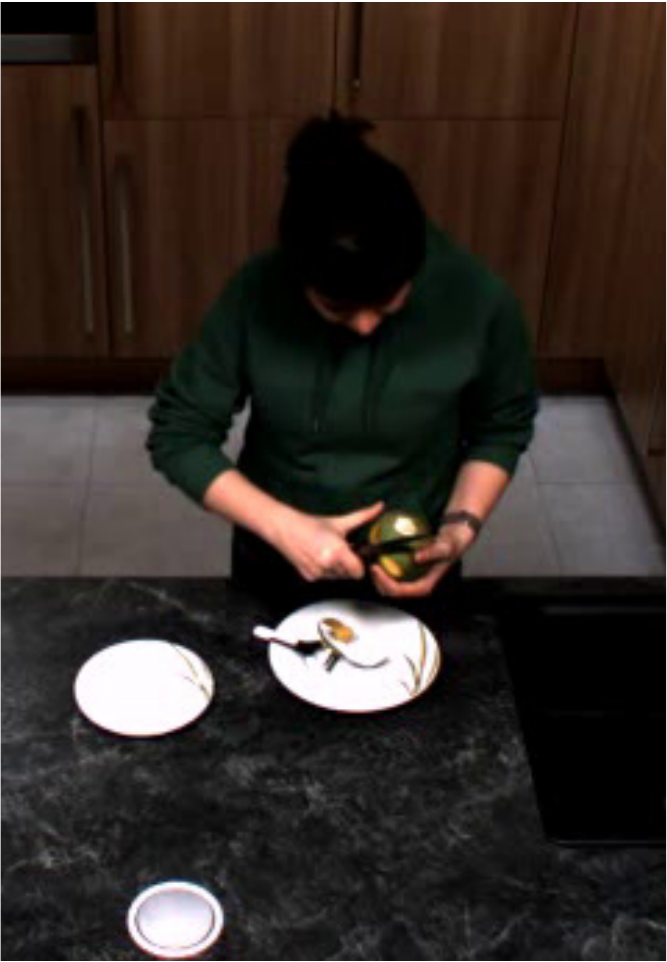}
          &\includegraphics[width=0.2\textwidth]{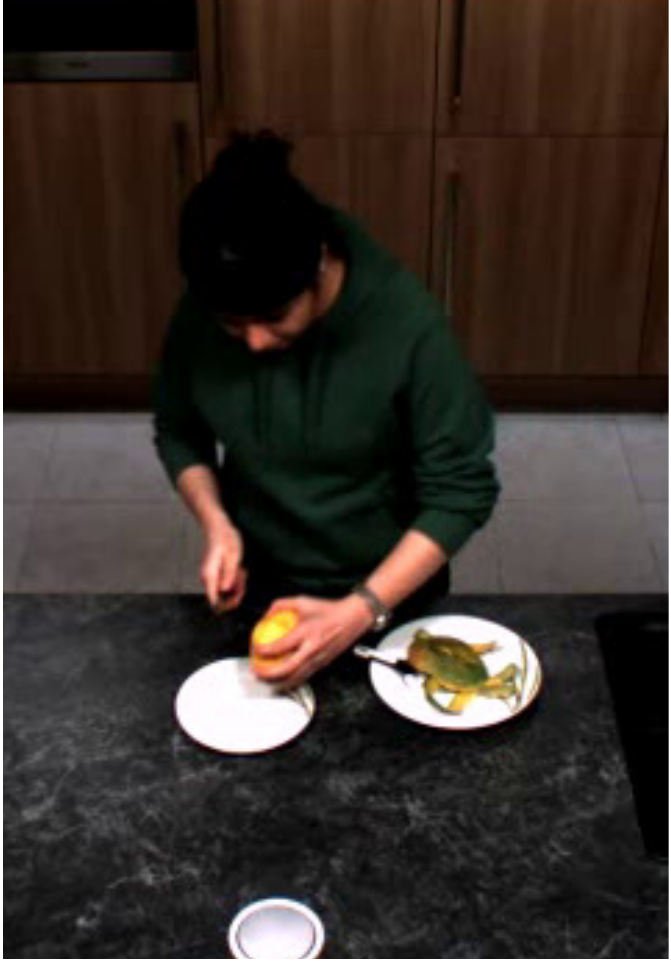}
          &\includegraphics[width=0.2\textwidth]{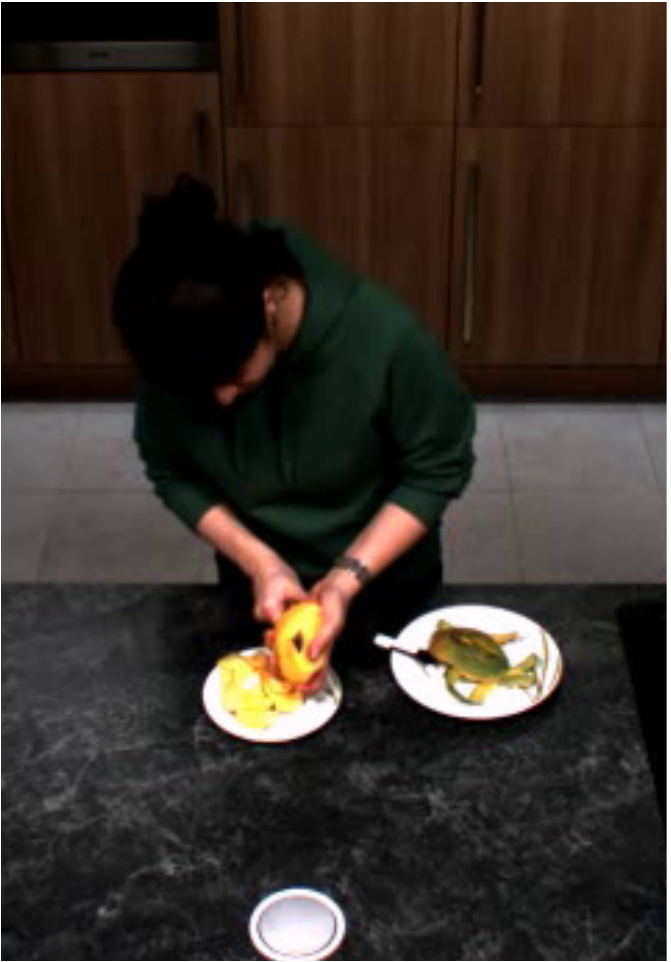}
          &\includegraphics[width=0.2\textwidth]{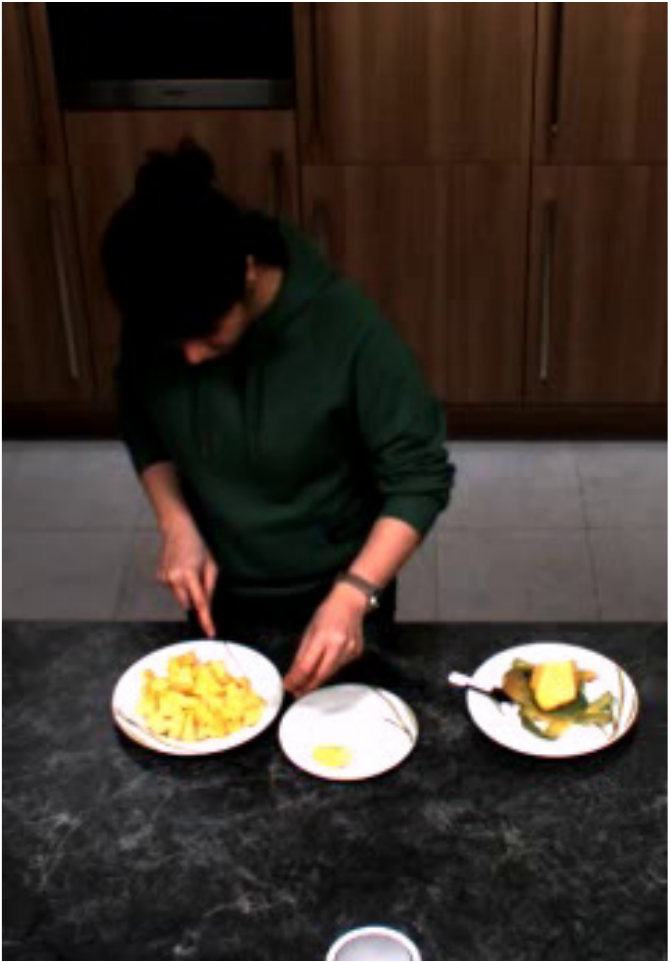}
          &\includegraphics[width=0.2\textwidth]{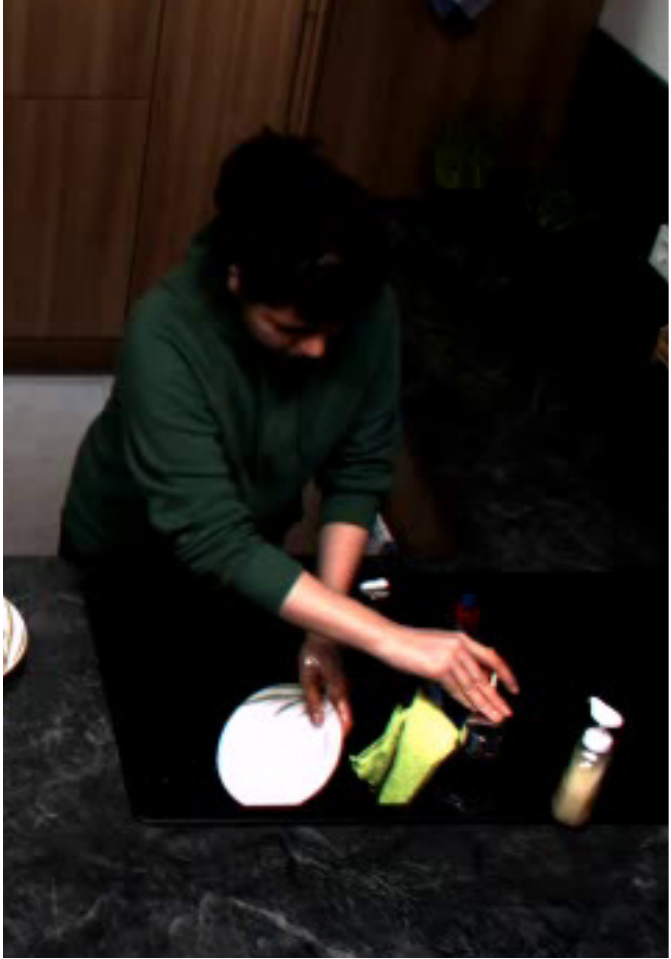}
        \end{tabular}\\      
        \begin{tabular}{l}
          \textit{\huge{The person peeled the fruit.}}\\
          \textit{\huge{The person put the fruit in the bowl.}}\\
          \textit{\huge{The person sliced the orange.}}\\
          \textit{\huge{The person put the pieces in the plate.}}\\
          \textit{\huge{The person rinsed the plate in the sink.}}\\
        \end{tabular}
      \end{tabular}
    \end{tabular}
  }    
  \vspace*{1ex}
  \caption{Some example sentences generated by our approach.
    The first row shows examples trained on YouTubeClips, where only
    one sentence is generated for each video.
    The second row shows examples trained on TACoS-MultiLevel, where
    paragraphs are generated.  
  }
  \label{fig:cherry-pick}
\end{figure*}

In this paper, we consider the problem of video captioning,
\ie\ generating one or multiple sentences to describe the content of a
video.
The given video could be as general as those uploaded to YouTube, or
it could be as specific as cooking videos with fine-grained activities.
This ability to generate linguistic descriptions for unconstrained
video is important because not only it is a critical step towards
machine intelligence, but also it has many applications in daily
scenarios such as video retrieval, automatic video subtitling, blind
navigation, \etc.
Figure~\ref{fig:cherry-pick} shows some example sentences generated by our
approach.

The video captioning problem has been studied for over one decade ever
since the first rule-based system on describing human activities with
natural language~\cite{Kojima2002}.
In a very limited setting, Kojima \etal designed some simple
heuristics for identifying video objects and a set of rules for
producing verbs and prepositions.
A sentence is then generated by filling predefined templates with the
recognized parts of speech.
Following their work, several succeeding approaches
\cite{Lee2008,Khan2011a,Khan2011b,Hanckmann2012,Barbu2012a} applied
similar rule-based systems to datasets with larger numbers of objects
and events, in different tasks and scenarios.
With \textit{ad hoc} rules, they manually establish the correspondence
between linguistic terms and visual elements, and analyze the
relations among the visual elements to generate sentences.
Among them, the most complex rule-based system~\cite{Barbu2012a}
supports a vocabulary of $118$ lexical entries (including $48$ verbs
and $24$ nouns).

To eliminate the tedious effort of rule engineering when the problem
scales, some recent methods train statistical models for lexical
entries, either in a fully
\cite{Das2013,Guadarrama2013,Krishnamoorthy2013,Sun2014} or weakly
\cite{Rohrbach2013,Rohrbach2014,Yu2015,XuR2015} supervised fashion.
The statistical models of different parts of speech usually have
different mathematical representations and training strategies (\eg,
\cite{Guadarrama2013,Krishnamoorthy2013}).
With most of the manual effort gone, the training process exposes
these methods to even larger datasets (\eg, YouTubeClips
\cite{Chen2011} and TACoS-MultiLevel \cite{Rohrbach2014}) which
contain thousands of lexical entries and dozens of hours of videos.
As a result, the video captioning task becomes much more challenging,
and the generation performance of these methods is usually low on
these large-scale datasets.

Since then, inspiring results have been achieved by a recent line of
work
\cite{Donahue2014,Venugopalan2014,Venugopalan2015,Pan2015,Xu2015,Yao2015}
which benefits from the rapid development of deep neural networks,
especially Recurrent Neural Network (RNN).
Applying RNN to translating visual sequence to natural language is
largely inspired by the recent advances in Neural Machine Translation
(NMT) \cite{Bahdanau2014,Sutskever2014} in the natural language
processing community.
The idea is to treat the image sequence of a video as the ``source
text'' and the corresponding caption as the target text.
Given a sequence of deep convolutional features (\eg,
VggNet~\cite{Simonyan14c} and C3D~\cite{Tran2015}) extracted from
video frames, a compact representation of the video is obtained by:
average pooling \cite{Venugopalan2014,Pan2015}, weighted average
pooling with an attention model \cite{Yao2015}, or taking the last
output from an RNN encoder which summarizes the feature sequence
\cite{Donahue2014,Venugopalan2015,Xu2015}.
Then an RNN decoder accepts this compact representation and outputs a
sentence of a variable length.

While promising results were achieved by these RNN methods, they only
focus on generating a single sentence for a short video clip.
So far the problem of generating multiple sentences or a paragraph for
a long video has not been attempted by deep learning approaches.
Some graphical-model methods, such as Rohrbach
\etal\cite{Rohrbach2014}, are able to generate multiple sentences, but
their results are still far from perfect.
The motivation of generating a paragraph is that most videos depict
far more than just one event.
Using only one short sentence to describe a semantically rich video
usually yields uninformative and even boring results.
For example, instead of saying \textit{the person sliced the potatoes,
  cut the onions into pieces, and put the onions and potatoes into the
  pot}, a method that is only able to produce one short sentence would
probably say \textit{the person is cooking}.

Inspired by the recent progress of document
modeling~\cite{Li2015,Lin2015} in natural language processing, we
propose a hierarchical-RNN framework for describing a long video with
a paragraph consisting of multiple sentences.
The idea behind our hierarchical framework is that we want to exploit
the temporal dependency among sentences in a paragraph, so that when
producing the paragraph, the sentences are not generated
independently.
Instead, the generation of one sentence might be affected by the
semantic context provided by the previous sentences. 
For example, in a video of cooking dishes, a sentence \textit{the
  person peeled the potatoes} is more likely to occur, than the
sentence \textit{the person turned on the stove}, after the sentence
\textit{the person took out some potatoes from the fridge}.
Towards this end, our hierarchical framework consists of two
generators, \ie\ a sentence generator and a paragraph generator, both
of which use recurrent layers for language modeling.
At the low level, the sentence generator produces single short
sentences that describe specific time intervals and video regions.
We exploit both temporal- and spatial-attention mechanisms to
selectively focus on visual elements when generating a sentence.
The embedding of the generated sentence is encoded by the output of
the recurrent layer.
At the high level, the paragraph generator takes the sentential
embedding as input, and uses another recurrent layer to output the
paragraph state, which is then used as the new initial state of the
sentence generator (see Section~\ref{sec:approach}).
Figure~\ref{fig:framework} illustrates our overall framework.
We evaluate our approach on two public datasets: YouTubeClips
\cite{Chen2011} and TACoS-MultiLevel \cite{Rohrbach2014}.
We show that our approach significantly outperforms other
state-of-the-art methods.
To our knowledge, this is the first application of hierarchical RNN to
video captioning task.

\section{Related Work}
\noindent\textbf{Neural Machine Translation.}\enspace The methods for
NMT
\cite{Kalchbrenner2013,Cho2014,Bahdanau2014,Sutskever2014,Li2015,Lin2015}
in computational linguistics generally follow the encoder-decoder
paradigm.
An encoder maps the source sentence to a fixed-length feature vector
in the embedding space.
A decoder then conditions on this vector to generate a translated
sentence in the target language.
On top of this paradigm, several improvements were proposed.
Bahdanau \etal\cite{Bahdanau2014} proposed a soft attention model to
do alignment during translation, so that their approach is able to
focus on different parts of the source sentence when generating
different translated words.
Li \etal\cite{Li2015} and Lin \etal\cite{Lin2015} employed
hierarchical RNN to model the hierarchy of a document.
Our approach is much similar to a neural machine translator with 
a simplified attention model and a hierarchical architecture.

\noindent\textbf{Image captioning with RNNs.}\enspace The first
attempt of visual-to-text translation using RNNs was seen in the work
of image captioning
\cite{Mao2015a,Kiros2014,Karpathy2014,Vinyals2014,Chen2014}, which can
be treated as a special case of video captioning when each video has a
single frame and no temporal structure.
As a result, image captioning only requires computing object
appearance features, but not action/motion features.
The amount of data handled by an image captioning method is much
(dozens of times) less than that handled by a video captioning method.
The overall structure of an image captioner (instance-to-sequence) is
also usually simpler than that of a video captioner
(sequence-to-sequence).
Some other methods, such as Park and Kim \cite{Park2015}, addressed
the problem of retrieving sentences from training database to describe
a sequence of images.
They proposed a local coherence model for fluent sentence transitions,
which serves a similar purpose of our paragraph generator.

\noindent\textbf{Video captioning with RNNs.}\enspace The very early
video captioning method~\cite{Venugopalan2014} based on RNNs extends
the image captioning methods by simply average pooling the video
frames.
Then the problem becomes exactly the same as image captioning.
However, this strategy works only for short video clips where there is
only one major event, usually appearing in one video shot from the
beginning to the end.
To avoid this issue, more sophisticated ways of encoding video
features were proposed in later work, using either a recurrent encoder
\cite{Donahue2014,Venugopalan2015,Xu2015} or an attention
model~\cite{Yao2015}.
Our sentence generator is closely related to Yao \etal\cite{Yao2015},
in that we also use attention mechanism to selectively focus on video
features.
One difference between our framework and theirs is that we
additionally exploit spatial attention.
The other difference is that after weighing video features with
attention weights, we do not condition the hidden state of our
recurrent layer on the weighted features (Section~\ref{sec:sent-gen}).

\section{Hierarchical RNN for Video Captioning}
\label{sec:approach}
Our approach stacks a paragraph generator on top of a sentence
generator.
The sentence generator is built upon 1) a Recurrent Neural Network
(RNN) for language modeling, 2) a multimodal layer~\cite{Mao2015a} for
integrating information from different sources, and 3) an attention
model~\cite{Yao2015,Bahdanau2014} for selectively focusing on the
input video features.
The paragraph generator is simply another RNN which models the
inter-sentence dependency.
It receives the compact sentential representation encoded by the
sentence generator, combines it with the paragraph history, and
outputs a new initial state for the sentence generator.
The RNNs exploited by the two generators incorporate the \emph{Gated
  Recurrent Unit} (GRU)~\cite{Cho2014} which is a simplification of
the Long Short-Term Memory (LSTM) architecture~\cite{Hochreiter1997}.
In the following, we first briefly review the RNN with the GRU (or the
\emph{gated RNN}), and then describe our framework in details.

\subsection{Gated Recurrent Unit}


A simple RNN~\cite{Elman1990} can be constructed by adding feedback
connections to a feedforward network that consists of three layers:
the input layer~$\mathbf{x}$, the hidden layer~$\mathbf{h}$, and the
output layer~$\mathbf{y}$.
The network is updated by both the input and the previous recurrent
hidden state as follows:
\begin{equation*}
  \begin{array}{l@{\hspace*{6ex}}r}
    \mathbf{h}^t =
    \phi\left(\mathbf{W}_h\mathbf{x}^t+\mathbf{U}_h\mathbf{h}^{t-1}+\mathbf{b}_h\right)
    &\text{(hidden state)}\\
    \mathbf{y}^t =
    \phi\left(\mathbf{U}_y\mathbf{h}^t+\mathbf{b}_y\right)
    &\text{(output)}\\
  \end{array}
\end{equation*}
where~$\mathbf{W},\mathbf{U}$ and~$\mathbf{b}$ are weight matrices and
biases to be learned, and~$\phi(\cdot)$ are element-wise activation
functions.

While the simple RNN is able to model temporal dependency for a
small time gap, it usually fails to capture long-term temporal
information.
To address this issue, the GRU~\cite{Cho2014} is designed to
adaptively remember and forget the past.
Inside the unit, the hidden state is modulated by non-linear gates.
%
Specifically, let $\odot$ denote the element-wise multiplication of
two vectors, the GRU computes the hidden state~$\mathbf{h}$ as:
\begin{equation*}
  \begin{array}{l@{\hspace*{-1ex}}r}
    \mathbf{r}^t=\sigma(\mathbf{W}_r\mathbf{x}^t+\mathbf{U}_r\mathbf{h}^{t-1}+\mathbf{b}_r)
    &\text{(reset gate)}\\
    \mathbf{z}^t=\sigma(\mathbf{W}_z\mathbf{x}^t+\mathbf{U}_z\mathbf{h}^{t-1}+\mathbf{b}_z)
    &\text{(update gate)}\\
    \widetilde{\mathbf{h}}^t=\phi\left(\mathbf{W}_h\mathbf{x}^t+\mathbf{U}_h(\mathbf{r}^t\odot\mathbf{h}^{t-1})+\mathbf{b}_h\right)\\
    \mathbf{h}^t=\mathbf{z}^t\odot\mathbf{h}^{t-1}+(1-\mathbf{z}^t)\odot\widetilde{\mathbf{h}}^t
    &\text{(hidden state)}\\
  \end{array}
\end{equation*}
where $\sigma(\cdot)$ are element-wise Sigmoid functions.
The reset gate~$\mathbf{r}$ determines whether the hidden state wants
to drop any information that will be irrelevant in the future.
The update gate~$\mathbf{z}$ controls how much information from the
previous hidden state will be preserved for the current state.
During the training of a gated RNN, the parameters can be estimated by
Backpropagation Through Time (BPTT)~\cite{Werbos1990} as in
traditional RNN architectures.

\begin{figure*}[t]
  \centering
  \includegraphics[width=0.9\textwidth]{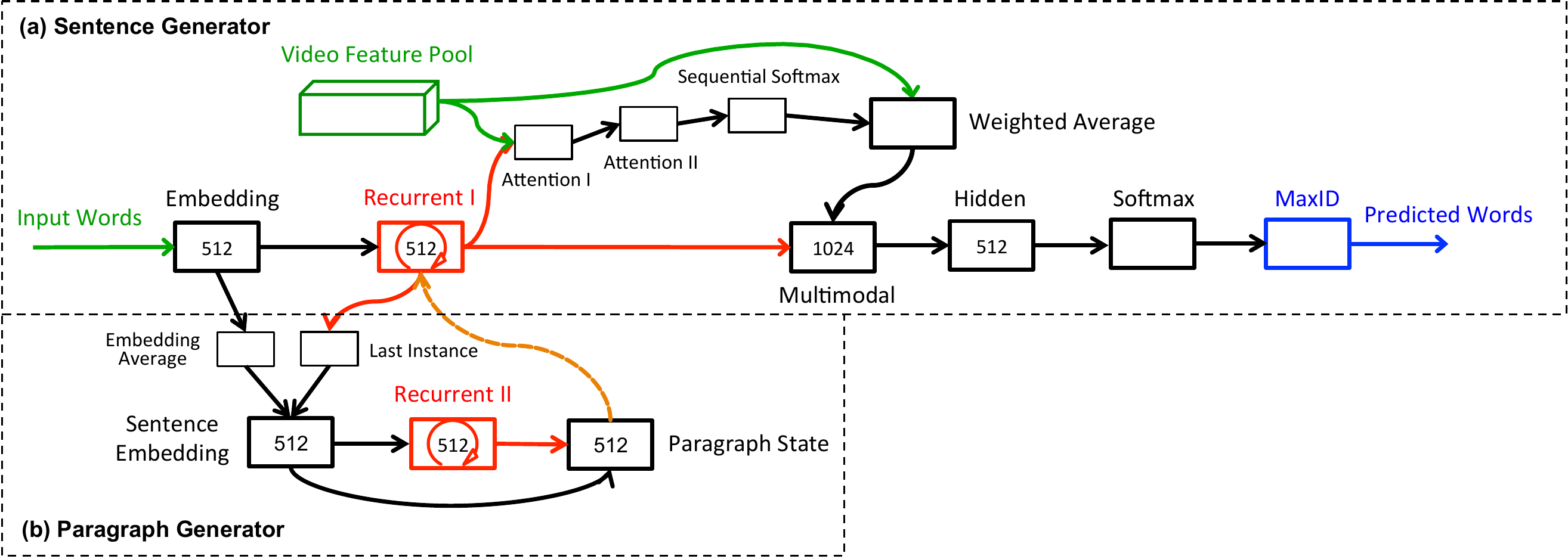}
  \caption{Our hierarchical RNN for video captioning.
    {\color{Green}Green} denotes the input to the framework,
    {\color{Blue}blue} denotes the output, and {\color{Red}red}
    denotes the recurrent components.
    The {\color{Orange}orange} arrow represents the
    reinitialization of the sentence generator with the current
    paragraph state.  
    For simplicity, we only draw a single video feature pool in the
    figure.
    In fact, both appearance and action features go through a
    similar attention process before they are fed into the multimodal
    layer.
  }
  \label{fig:framework}
\end{figure*}

\subsection{Sentence Generator}
\label{sec:sent-gen}
The overall structure of our hierarchical RNN is illustrated in
Figure~\ref{fig:framework}.
The sentence generator operates at every time step when a one-hot
input (1-of-$N$ encoding, where $N$ is the vocabulary size) arrives
at the embedding layer.
The embedding layer converts the one-hot vector to a dense
representation in a lower dimensional space by multiplying it with an
embedding table ($512\times N$), of which each row is a word embedding
to be learned.
The resulting word embedding is then input to our first RNN, \ie, the
recurrent layer I.
This gated recurrent layer has $512$ dimensions and acts similarly to
those that are commonly employed by a variety of image/video
captioning methods (\eg, \cite{Venugopalan2015,Mao2015a,Yao2015}),
\ie, modeling the syntax of a language.
It updates its hidden state every time a new word arrives, and encodes
the sentence semantics in a compact form up to the words that have
been fed in.
We set the activation function~$\phi$ of this recurrent layer to be
the Rectified Linear Unit (ReLU)~\cite{Nair2010}, since it performs
better than non-linear activation functions such as Sigmoid according
to our observation.

As one branch, the output of the recurrent layer I is directed to the
attention layers to compute attention weights for the features in the
video feature pool.
Our attention model is inspired by the recent soft-alignment method
that has been successfully applied in the context of Neural
Machine Translation (NMT)~\cite{Bahdanau2014}, and was later adapted
to video captioning by Yao \etal\cite{Yao2015}.
The difference between our model and the one used by Yao \etal is that
their model only focuses on temporal attention.
We additionally include spatial attention by computing features for
multiple image patches at different locations on a video frame and
pool the features together.
This simple improvement is important when objects are small and
difficult to be localized on some datasets
(\eg, TACoS-MultiLevel~\cite{Rohrbach2014}).
In this case, whole-frame-based video features will fail to capture
the object information and multiple object proposals are needed for
good performance (see Section~\ref{sec:experiments} for details).
Let the features in the pool be denoted as
$\{\mathbf{v}_1,\mathbf{v}_2,\ldots,\mathbf{v}_{KM}\}$, where $M$ is
the video length and $K$ is the number of patches on each frame.
We want to compute a set of weights
$\{\beta_1^t,\beta_2^t,\ldots,\beta_{KM}^t\}$ for these features at
each time step~$t$ such that $\sum_{m=1}^{KM}\beta_m^t = 1$.
To do so, we first compute an attention score~$q_m^t$ for each
frame~$m$, conditioning on the previous hidden
state~$\mathbf{h}^{t-1}$:
\[q_m^t=\mathbf{w}^{\top}\phi(\mathbf{W}_q\mathbf{v}_m+\mathbf{U}_q\mathbf{h}^{t-1}+\mathbf{b}_q)\]
where $\mathbf{w}$, $\mathbf{W}_q$, $\mathbf{U}_q$, and~$\mathbf{b}_q$
are the parameters shared by all the features at all the time steps,
and~$\phi$ is set to the element-wise Scaled Hyperbolic Tangent (stanh)
function~\cite{LeCun1998}: $1.7159\cdot\text{tanh}(\frac{2x}{3})$.
The above computation is performed by the attention layers I and II in
Figure~\ref{fig:framework}(a), where the attention layer I projects
the feature~$\mathbf{v}$ and the hidden state~$\mathbf{h}$ into a
lower dimensional space whose dimension can range from $32$ to $256$.
The attention layer II then further compresses the activation of the
projected vector into a scalar, one for each feature.
After this, we set up a sequential softmax layer to get the
attention weights:
\[\beta_m^t=\exp\left(q_m^t\right) \Big/ \sum_{m'=1}^{KM}\exp\left(q_{m'}^t\right) \]
Finally, a single feature vector is obtained by weighted averaging:
$\mathbf{u}^t=\sum_{m=1}^{KM}\beta_m^t\mathbf{v}_m$.
The above process is a sophisticated version of the temporal mean
pooling.
It allows the sentence generator to selectively focus on a
subset of the features during generation.
Note that while only one feature channel is shown in
Figure~\ref{fig:framework}(a), our sentence generator in fact pumps
features of several channels through the same attention process.
Each feature channel has a different set of weights and biases to be
learned.
In our experiments, we employ two feature channels, one for object
appearance and the other for action/motion.
(Section~\ref{sec:experiments}).

After the attention process, the weighted sums of the video features
are fed into the multimodal layer which has $1024$ dimensions.
The multimodal layer also receives the output of the recurrent layer
I, thus connecting the vision component with the language model.
Suppose we have two video feature channels, of which the weighted
features output by the attention model are~$\mathbf{u}_o^t$
and~$\mathbf{u}_a^t$ respectively.
The multimodal layer maps the two features, together with the hidden
state~$\mathbf{h}^t$ of the recurrent layer I, into a $1024$
dimensional feature space and add them up:
\[\mathbf{m}^t=\phi(\mathbf{W}_{m,o}\mathbf{u}_o^t+\mathbf{W}_{m,a}\mathbf{u}_a^t+\mathbf{U}_m\mathbf{h}^t+\mathbf{b}_m)\]
where~$\phi$ is set to the element-wise stanh function.
To reduce overfitting, we add dropout~\cite{Srivastava2014} with a
drop rate of 0.5 to this layer.

The multimodal layer is followed by a hidden layer and a softmax layer
(see Figure~\ref{fig:framework}(a)), both with the element-wise stanh
function as their activation functions.
The hidden layer has exactly the same dimension $512$ with the word
embedding layer, and the softmax layer has a dimension that is equal
to the size of the vocabulary which is dataset-dependent.
Inspired by the transposed weight sharing scheme recently proposed by
Mao \etal\cite{Mao2015b}, we set the projection matrix from the
hidden layer to the softmax layer as the transpose of the word
embedding table.
It has been shown that this strategy allows the use of a word
embedding layer with a much larger dimension due to the parameter
sharing, and helps regularize the word embedding table because of the
matrix transpose.
As the final step of the sentence generator, the maxid layer picks the
index that points to the maximal value in the output of the softmax
layer.
The index is then treated as the predicted word id.
Note that during test, the predicted word will be fed back to the
sentence generator again as the next input word.
While in the training, the next input word is always provided by the
annotated sentence.

\subsection{Paragraph Generator}
The sentence generator above only handles one single sentence at a
time.
For the first sentence in the paragraph, the initial state of the
recurrent layer I is set to all zeros, \ie, $\mathbf{h}^0=\mathbf{0}$.
However, any sentence after that will have its initial state
conditioned on the semantic context of all its preceding sentences.
This semantic context is encoded by our paragraph generator.

During the generation of a sentence, an embedding average layer (see
Figure~\ref{fig:framework}(b)) accumulates all the word embeddings of
the sentence and takes the average to get a compact embedding vector.
The average strategy is inspired by the QA embedding~\cite{Bordes2014}
in which questions and answers are both represented as a combination
of the embeddings of their individual words and/or symbols.
We also take the last state of the recurrent layer I as a compact
representation for the sentence, following the idea behind the
Encoder-Decoder framework~\cite{Cho2014} in NMT.
After that, the averaged embedding and the last recurrent state are
concatenated together, and fully connected to the sentence embedding
layer ($512$ dimensions) with stanh as the activation function.
We treat the output of the sentence embedding layer as the final
sentence representation.

The sentence embedding layer is linked to our second gated RNN (see
Figure~\ref{fig:framework}(b)).
The recurrent layer II operates whenever a full sentence goes through
the sentence generator and the sentence embedding is produced by the
sentence embedding layer.
Thus the two recurrent layers are asynchronous: while the recurrent
layer I keeps updating its hidden state at every time step, the
recurrent layer II only updates its hidden state when a full sentence
has been processed.
The recurrent layer II encodes the paragraph semantics in a compact
form up to the sentences that have been fed in.
Finally, we set up a paragraph state layer to combine the hidden state
of the recurrent layer II and the sentence embedding.
This paragraph state is used as the initial hidden state when the
recurrent layer I is reinitialized for the next sentence.
It essentially provides the sentence generator with the paragraph
history so that the next sentence is produced in the context.

\section{Training and Generation}
\label{sec:tg}
We train all the components in our hierarchical framework together
from scratch with randomly initialized parameters.
We treat the activation value indexed by a training word~$w_t^n$ in
the softmax layer of our sentence generator as the likelihood of
generating that word: 
\[P\left(w_t^n|s_{1:n-1},w_{1:t-1}^n,\mathbf{V}\right)\]
given 1) all the preceding sentences $s_{1:n-1}$ in the paragraph, 2)
all the previous words $w_{1:t-1}^n$ in the same sentence~$n$, and 3)
the corresponding video~$\mathbf{V}$.
The cost of generating that training word is then defined as the
negative logarithm of the likelihood.
We further define the cost of generating the whole paragraph $s_{1:N}$
($N$ is the number of sentences in the paragraph) as:
\begin{equation*}
  \begin{array}{l}
  \mathcal{PPL}(s_{1:N}|\mathbf{V})\\
  =-\displaystyle\sum_{n=1}^N\sum_{t=1}^{T_n}
  \log P\left(w_t^n|s_{1:n-1},w_{1:t-1}^n,\mathbf{V}\right)
  \Bigg/ \displaystyle\sum_{n=1}^NT_n\\
  \end{array}
\end{equation*}
where~$T_n$ is the number of words in the sentence~$n$.
The above cost is in fact the \emph{perplexity} of the paragraph given
the video.
Finally, the cost function over the entire training set is defined as:
\begin{equation}
  \label{eq:ppl}
  \mathcal{PPL}=\displaystyle\sum_{y=1}^Y\left(
  \mathcal{PPL}(s_{1:N_y}^y|\mathbf{V}_y)\cdot\sum_{n=1}^{N_y}T_n^y\right)
  \Bigg/ {\displaystyle\sum_{y=1}^Y\sum_{n=1}^{N_y}T_{n}^y}
\end{equation}
where~$Y$ is the total number of paragraphs in the training set.
To reduce overfitting, L2 and L1 regularization terms are added
to the above cost function.
We use Backpropagation Through Time (BPTT)~\cite{Werbos1990} to
compute the gradients of the parameters and Stochastic Gradient
Descent (SGD) to find the optimum.
For better convergence, we divide the gradient by a running average
of its recent magnitude according to the RMSPROP
algorithm~\cite{Tieleman2012}.
We set a small learning rate $10^{-4}$ to avoid the gradient
explosion problem that is common in the training process of RNNs.

After the parameters are learned, we perform the generation with Beam
Search.
Suppose that we use a beam width of $L$.
The beam search process starts with the BOS (begin-of-sentence)
symbol~$w_{\text{BOS}}$ (\ie, $w_0$) which is treated as a 1-word
sequence with zero cost at~$t=0$.
Assume that at any time step~$t$, there are at most~$L$ $t$-word
sequences that were previously selected with the lowest sequence costs
(a sequence cost is the sum of the word costs in that sequence).
For each of the $t$-word sequences, given its last word as input, the
sentence generator calculates the cost of the next word $-\log
P(w_t|w_{1:t-1},\mathbf{V})$ and the sequence cost if the word is
appended to the sequence.
Then from all the $t+1$-word sequences expanded from the existing
$t$-word sequences, we pick the top~$L$ with the lowest sequence
costs.

Of the new $t+1$-word sequences, any one that is a complete sentence
(\ie, the last word~$w_{t+1}$ is the EOS (end-of-sentence)
symbol~$w_{\text{EOS}}$) will be removed from the search tree.
It will be put into our sentence pool if 1) there are less than~$J$
($J\le L$) sentences in the pool or, 2) its sequence cost is lower
than one of the~$J$ sentences in the pool.
In the second case, the sentence with the highest cost will be removed
from the pool, replaced by the new added sentence.
Also of the new $t+1$-word sequences, any one that has a higher
sequence cost than all of the~$J$ sentences in the pool will be
removed from the search tree.
The reason is that expanding a word sequence monotonically increases
its cost.
The beam search process stops when there is no word sequence to be
expanded in the next time step.
In the end, $J$ candidate sentences will be generated for
post-processing and evaluation.

After this, the generation process goes on by picking the sentence
with the lowest cost from the~$J$ candidate sentences.
This sentence is fed into our paragraph generator which reinitializes
the sentence generator.
The sentence generator then accepts a new BOS and again produces~$J$
candidate sentences.
This whole process stops when the sentence received by the paragraph
generator is the EOP (end-of-paragraph) which consists of only the BOS
and the EOS.
Finally, we will have a paragraph that is a sequence of lists, each
list with ~$J$ sentences.
In our experiments, we set $L=J=5$.
Excluding the calculation of visual features, the average
computational time for the sentence generator to produce top 5
candidate sentences with a beam width of 5 is 0.15 seconds, on a
single thread with CPU Intel(R) Core(TM) i7-5960X @ 3.00GHz.

\section{Experiments}
\label{sec:experiments}
We evaluate our approach on two benchmark datasets:
YouTubeClips~\cite{Chen2011} and TACoS-MultiLevel~\cite{Rohrbach2014}.

\noindent\textbf{YouTubeClips}\enspace This dataset consists of
$1,967$ short video clips ($9$ seconds on average) downloaded from
YouTube.
The video clips are open-domain, containing different people,
animals, actions, scenarios, landscapes, \etc.
Each video clip is annotated with multiple parallel sentences by
different turkers.
There are $80,839$ sentences in total, with about $41$ annotated
sentences per clip.
Each sentence on average contains about $8$ words.
The words contained in all the sentences constitute a vocabulary of
$12,766$ unique lexical entries.
We adopt the train and test splits provided by Guadarrama
\etal\cite{Guadarrama2013}, where $1,297$ and $670$ videos are used
for training and testing respectively.
It should be noted that while multiple sentences are annotated for
each video clip, they are \emph{parallel} and \emph{independent} in
the temporal extent, \ie, the sentences describe exactly the same
video interval, from the beginning to the end of the video.
As a result, we use this dataset as a special test case for our
approach, when the paragraph length $N=1$.

\noindent\textbf{TACoS-MultiLevel}\enspace This dataset consists of
$185$ long videos (6 minutes on average) filmed in an indoor
environment.
The videos are closed-domain, containing different actors,
fine-grained activities, and small interacting objects in daily
cooking scenarios.
Each video is annotated by multiple turkers.
A turker annotates a sequence of temporal intervals across the video,
pairing every interval with a single short sentence.
There are $16,145$ distinct intervals and $52,478$ sentences in total,
with about $87$ intervals and $284$ sentences per video.
The sentences were originally preprocessed so that they all have the
past tense, and different gender specific identifiers were substituted
with ``\textit{the person}''.
Each sentence on average contains about $8$ words.
The words contained in all the sentences constitute a vocabulary of
$2,864$ unique lexical entries.
We adopt the train and test splits used by Rohrbach~\etal
\cite{Rohrbach2014}, where $143$ and $42$ videos are used for training
and testing respectively.
Note that the cooking activities in this dataset have strong temporal
dependencies.
Such dependency in a video is implied by the sequence of intervals
annotated by the same turker on that video.
Following Donahue~\etal \cite{Donahue2014} and Rohrbach~\etal
\cite{Rohrbach2014}, we employ the interval information to align our
sentences in the paragraph during both training and generation.
This dataset is used as a general test case for our approach, when the
paragraph length $N>1$.

To model video object appearance, we use the pretrained
VggNet~\cite{Simonyan14c} (on the ImageNet
dataset~\cite{Russakovsky2015}) for both datasets.
Since the objects in YouTubeClips are usually prominent, we only
extract one VggNet feature for each entire frame.
This results in only temporal attention in our sentence generator
(\ie, $K=1$ in Section~\ref{sec:sent-gen}).
For TACoS-MultiLevel, the interacting objects are usually quite small
and difficult to be localized.
To solve this problem, both Donahue~\etal \cite{Donahue2014} and
Rohrbach~\etal \cite{Rohrbach2014} designed a specialized hand
detector.
Once the hand regions are detected, they extract features in the
neighborhood to represent the interacting objects.
Instead of trying to accurately locate hands which requires a lot of
engineering effort as in their case, we rely on a simple routine to
obtain multiple object proposals.
We first use Optical Flow~\cite{Farneback2003} to roughly detect
a bounding box for the actor in each frame.
We then extract $K$ image patches of size $220\times220$ along the
lower part of the box border, where every two neighboring patches have
an overlap of half their size.
Our simple observation is that these patches together have a high
recall of containing the interacting objects while the actor is
cooking.
Finally, we compute the VggNet feature for each patch and pool all the
patch features.
When $K>1$, the above routine leads to both temporal and spatial
attention in our sentence generator.
In practice, we find that a small value of $K$ (\eg, $3\sim5$) is enough to
yield good performance.

To model video motion and activities, we use the pretrained
C3D~\cite{Tran2015} (on the Sports-1M dataset~\cite{Karpathy2014}) for
YouTubeClips.
The C3D net reads in a video and outputs a fixed-length feature vector
every $16$ frames.
Thus when applying the attention model to the C3D feature pool, we set
$K=1$ and divide $M$ by $16$ (Section~\ref{sec:sent-gen}).
For the TACoS-MultiLevel dataset, since the cooking activities are
fine-grained, the same model trained on sports videos does not work
well.
Alternatively we compute the Dense Trajectories~\cite{Wang2011} for
each video interval and encode them with the Fisher
vector~\cite{Jegou2012}.
For the attention model, we set $K=1$ and $M=1$.

We employ three different evaluation metrics:
BLEU \cite{Papineni2002}, METEOR \cite{Banerjee2005}, and
CIDEr \cite{Vedantam2014}.
Because the YouTubeClips dataset was tested on by most existing
video-captioning methods, the prior results of all the three metrics
have been reported.
The TACoS-MultiLevel dataset is relatively new and only the BLEU scores
were reported in the previous work.
We compute the other metrics for the comparison methods based on the
generated sentences that come with the dataset.
Generally, the higher the metric scores are, the better the generated
sentence correlates with human judgment.
We use the evaluation script provided by Chen \etal\cite{Chen2015} to
compute scores on both datasets.

\begin{table}[t]
  \centering
  \resizebox{0.49\textwidth}{!}{
    \begin{tabular}{@{}l|cccc|c|c}
      & \textbf{B@1} & \textbf{B@2} & \textbf{B@3} & \textbf{B@4} &
      \textbf{M} & \textbf{C}\\
      \hline
      \hline
      LSTM-YT~\cite{Venugopalan2014}&
      - & - & - & 0.333 & 0.291 & -\\
      S2VT~\cite{Venugopalan2015}&
      - & - & - & - & 0.298 & -\\
      MM-VDN~\cite{Xu2015}&
      - & - & - & 0.376 & 0.290 & -\\
      TA~\cite{Yao2015}&
      0.800 & 0.647 & 0.526 & 0.419 & 0.296 & 0.517\\
      LSTM-E~\cite{Pan2015}&
      0.788 & 0.660 & 0.554 & 0.453 & 0.310 & -\\
      \hline
      h-RNN-Vgg&
      0.773 & 0.645 & 0.546 & 0.443 & 0.311 & 0.621\\
      h-RNN-C3D&
      0.797 & 0.679 & 0.579 & 0.474 & 0.303 & 0.536 \\
      h-RNN (Ours)&
      \textbf{0.815} & \textbf{0.704} & \textbf{0.604} 
      & \textbf{0.499} & \textbf{0.326} & \textbf{0.658} \\
    \end{tabular}
  }
  \vspace*{1ex}  
  \caption{Results on YouTubeClips, where B, M, and C are
    short for BLEU, METEOR, and CIDEr respectively.  
  }
  \label{tab:youtube}
\end{table}

\subsection{Results}
We compare our approach (h-RNN) on YouTubeClips with six
state-of-the-art methods: LSTM-YT~\cite{Venugopalan2014},
S2VT~\cite{Venugopalan2015}, MM-VDN~\cite{Xu2015}, TA~\cite{Yao2015},
and LSTM-E~\cite{Pan2015}.
Note that in this experiment a single sentence is generated for each
video.
Thus only our sentence generator is evaluated in comparison to others.
To evaluate the importance of our video features, we also report the
results of two baseline methods: h-RNN-Vgg and h-RNN-C3D.
The former uses only the object appearance feature and the latter uses
only the motion feature, with other components of our framework unchanged.
The evaluation results are shown in Table~\ref{tab:youtube}.
We can see that our approach performs much better than the comparison
methods, in all the three metrics.
The improvements on the most recent state-of-the-art method (\ie,
LSTM-E \cite{Pan2015}) are $\frac{0.499-0.453}{0.453}=10.15\%$ in the
BLEU@4 score, and $\frac{0.326-0.310}{0.310}=5.16\%$ in the METEOR
score.
Since LSTM-E also exploits VggNet and C3D features, this demonstrates
that our sentence generator framework is superior to their joint
embedding framework.
Moreover, although TA \cite{Yao2015} also employs temporal
attention, our approach produces much better results due to the fact
that the hidden state of our RNN is not conditioned on the video
features.
Instead, the video features are directly input to our multimodal
layer.
%
Our approach also outperforms the two baseline methods by large
margins, indicating that both video features are indeed crucial in the
video captioning task.

We compare our approach on TACoS-MultiLevel with three
state-of-the-art methods: CRF-T~\cite{Rohrbach2013},
CRF-M~\cite{Rohrbach2014}, and LRCN~\cite{Donahue2014}.
Like above, we have two baseline methods h-RNN-Vgg and h-RNN-DT which
use only the appearance and motion features respectively.
We also add another two baseline methods RNN-sent and RNN-cat that
have no hierarchy (\ie, with only the sentence generator, but not the
paragraph generator).
RNN-sent is trained and tested on individual video clips that are
segmented from the original $185$ long videos according to the
annotated intervals.
The initial state of the sentence generator is set to zero for each
sentence.
As a result, sentences are trained and generated independently.
RNN-cat initializes the sentence generator with zero only for the
first sentence in a paragraph.
Then the sentence generator maintains its state for the following
sentences until the end of the paragraph.
This concatenation strategy for training a paragraph has been
exploited in a recent neural conversational model~\cite{Vinyals2015}.
We use RNN-send and RNN-cat to evaluate the importance of our
hierarchical structure.

\begin{table}[t]
  \centering
  \resizebox{0.49\textwidth}{!}{
  \begin{tabular}{@{}c}
    \begin{tabular}{@{}l|cccc|c|c@{}}
      & \textbf{B@1} & \textbf{B@2} & \textbf{B@3} & \textbf{B@4} &
      \textbf{M} & \textbf{C}\\
      \hline
      \hline
      CRF-T~\cite{Rohrbach2013} & 0.564 & 0.447 & 0.332 & 0.253 &
      0.260 & 1.248\\
      CRF-M~\cite{Rohrbach2014} & 0.584 & 0.467 & 0.352 & 0.273 &
      0.272 & 1.347\\
      LRCN~\cite{Donahue2014} & 0.593 & 0.482 & 0.370 & 0.292 & 0.282
      & 1.534\\
      \hline
      h-RNN-Vgg & 0.561 & 0.445 & 0.329 & 0.256 & 0.260 & 1.267\\
      h-RNN-DT & 0.557 & 0.451 & 0.346 & 0.274 & 0.261 & 1.400\\
      RNN-sent & 0.568 & 0.469 & 0.367 & 0.295 & 0.278 & 1.580\\
      RNN-cat & 0.605 & 0.489 & 0.376 & 0.297 & 0.284 & 1.555\\
      h-RNN (Ours) & \textbf{0.608} & \textbf{0.496} & \textbf{0.385} &
      \textbf{0.305} & \textbf{0.287} & \textbf{1.602}\\
    \end{tabular}
  \end{tabular}
  }
  \vspace*{1ex}  
  \caption{Results on TACoS-MultiLevel, where B, M, and C are
    short for BLEU, METEOR, and CIDEr respectively. 
    %
  }
  \label{tab:tacos}
\end{table}

\begin{figure}[t]
  \centering
  \resizebox{0.41\textwidth}{!}{
    \begin{tabular}{l}
      \begin{tabular}{c@{\hspace{1ex}}c@{\hspace{1ex}}c@{\hspace{1ex}}c}
        \includegraphics[width=0.15\textwidth]{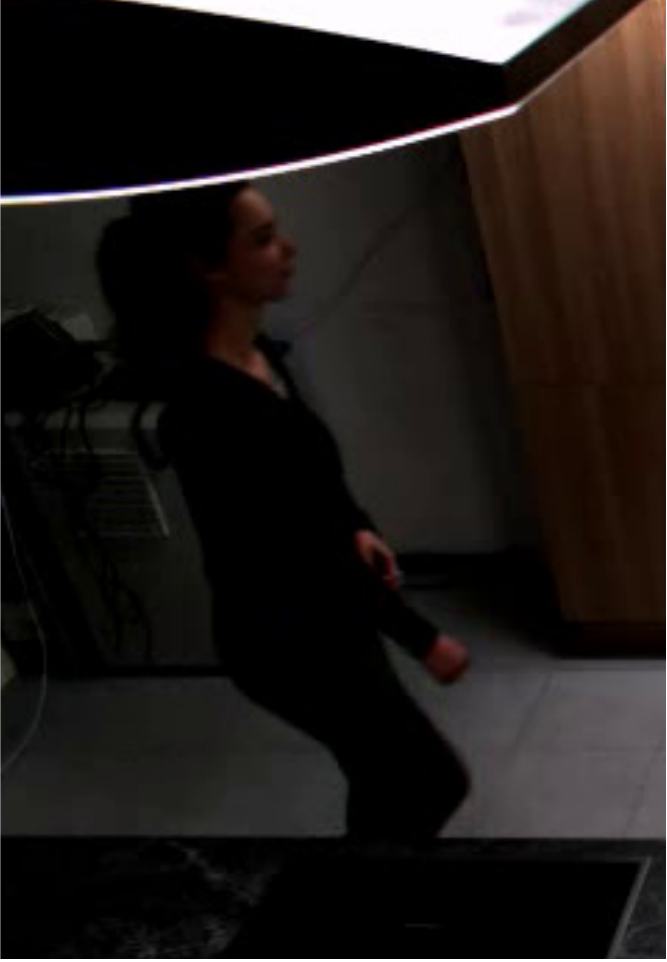}
        &\includegraphics[width=0.15\textwidth]{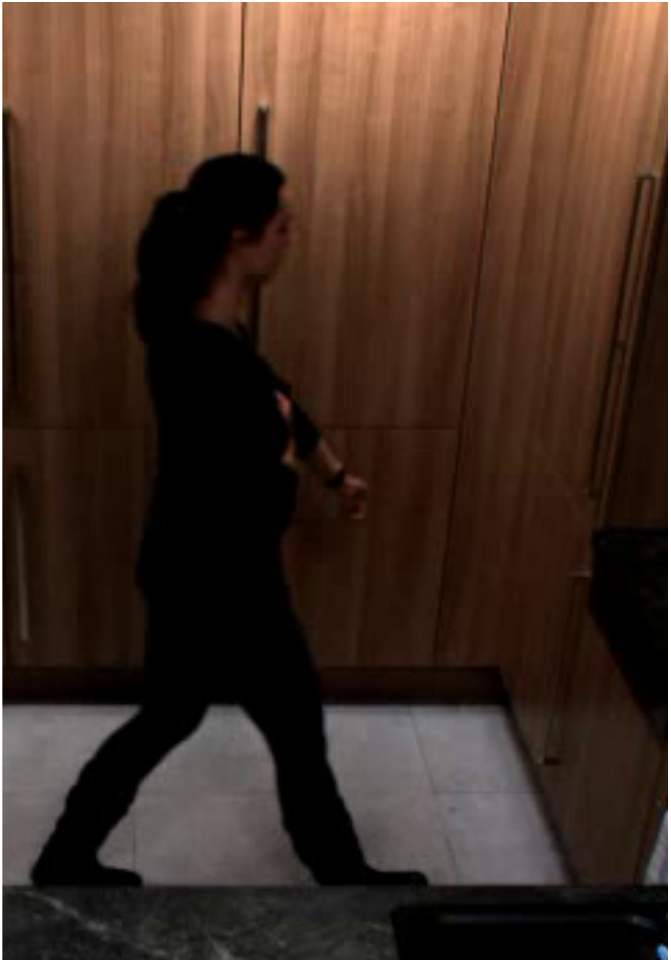}
        &\includegraphics[width=0.15\textwidth]{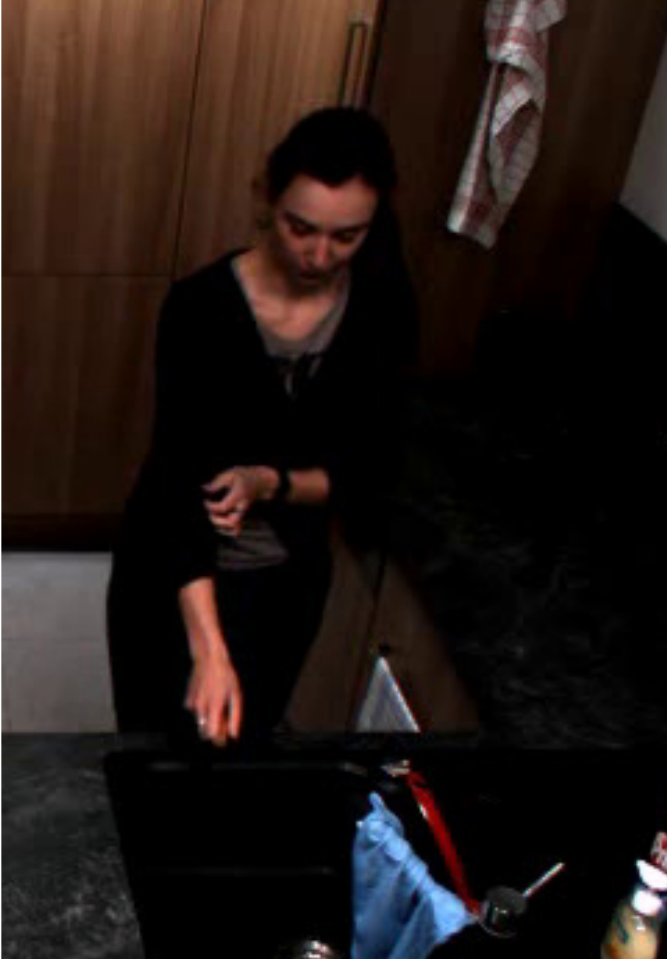}
        &\includegraphics[width=0.15\textwidth]{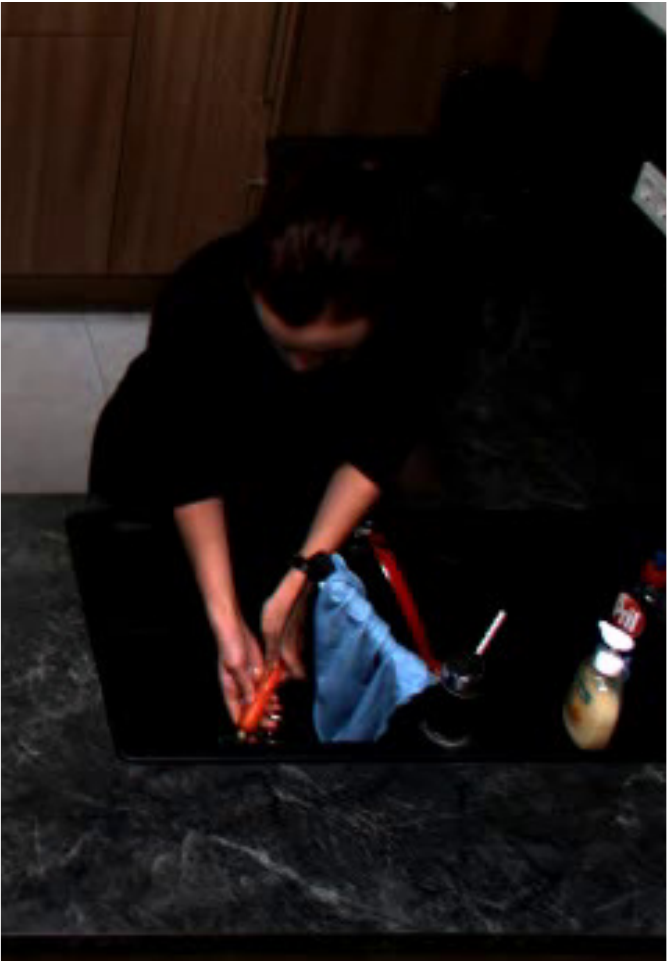}\\
      \end{tabular}\\
      \begin{tabular}{l|l}
        RNN-sent:
        &
        \begin{tabular}{l}
          \textit{The person entered the kitchen.}\\
          \textit{The person went to the refrigerator.}\\
          {\color{Red}\textit{The person placed the cucumber on the
              cutting board.}}\\
          {\color{Red}\textit{The person rinsed the
              cutting board.}}\\
        \end{tabular}\\
        \hline
        h-RNN:
        &
        \begin{tabular}{l}
          \textit{The person walked into the kitchen.}\\
          \textit{The person went to the refrigerator.}\\
          {\color{Green}\textit{The person walked over
              to the sink.}}\\
          {\color{Green}\textit{The person rinsed the
              carrot in the sink.}}\\
        \end{tabular}
      \end{tabular}
      \\\\
      \begin{tabular}{c@{\hspace{1ex}}c@{\hspace{1ex}}c@{\hspace{1ex}}c}      
        \includegraphics[width=0.15\textwidth]{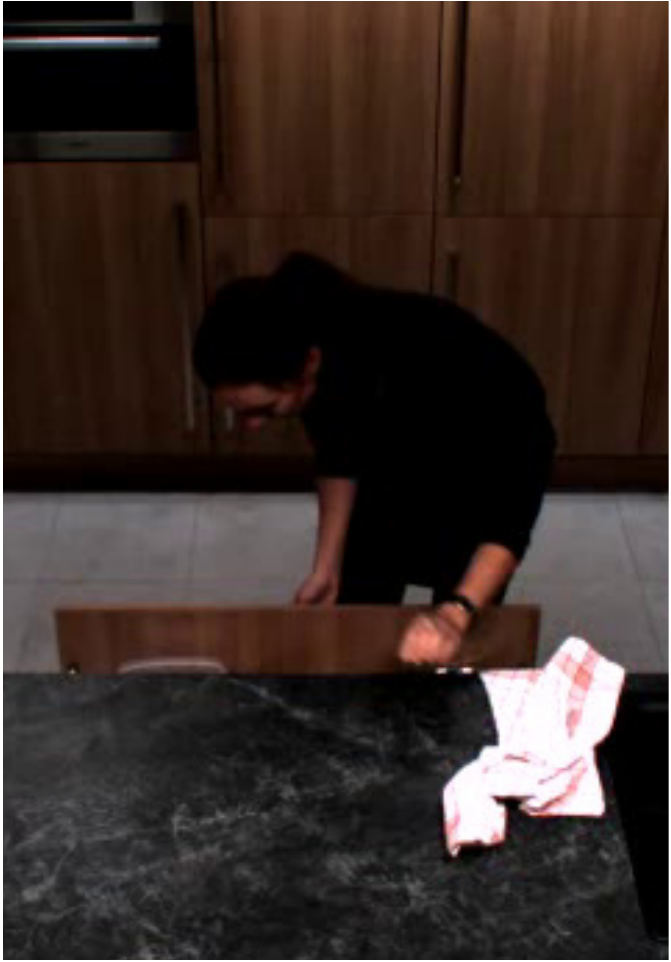}
        &\includegraphics[width=0.15\textwidth]{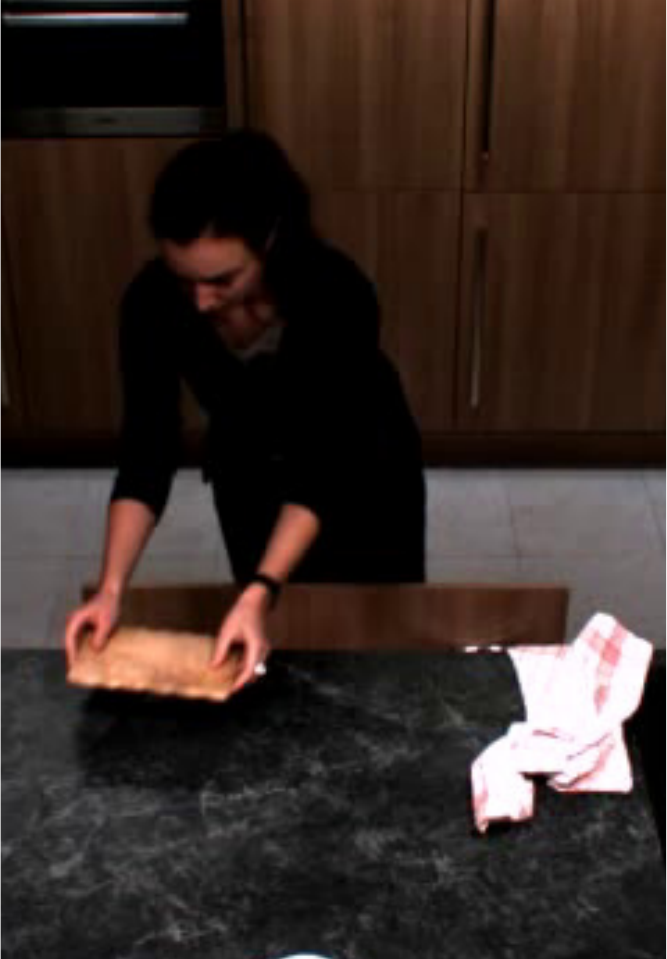}
        &\includegraphics[width=0.15\textwidth]{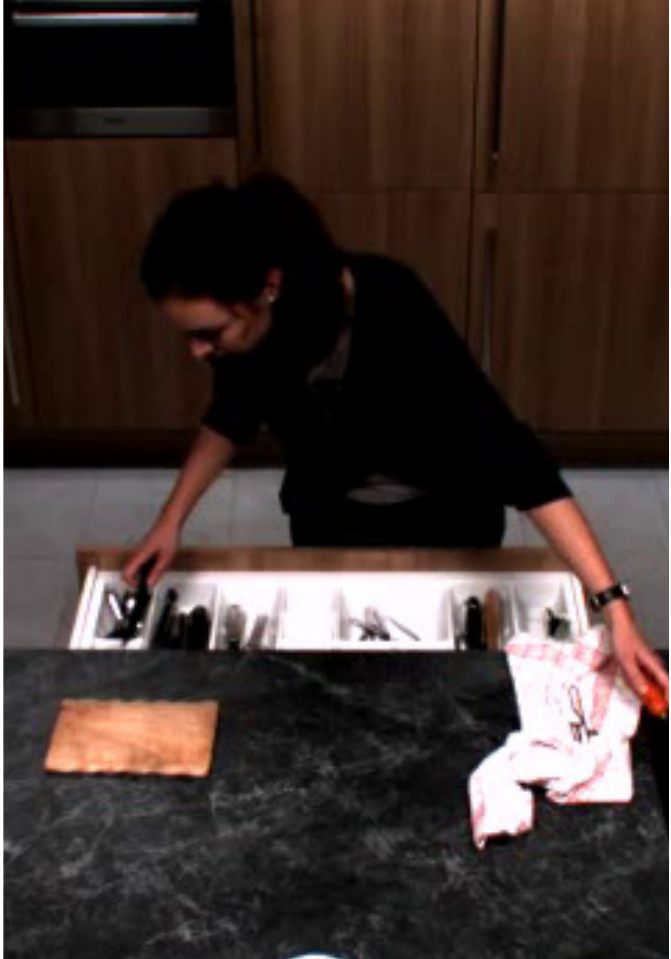}
        &\includegraphics[width=0.15\textwidth]{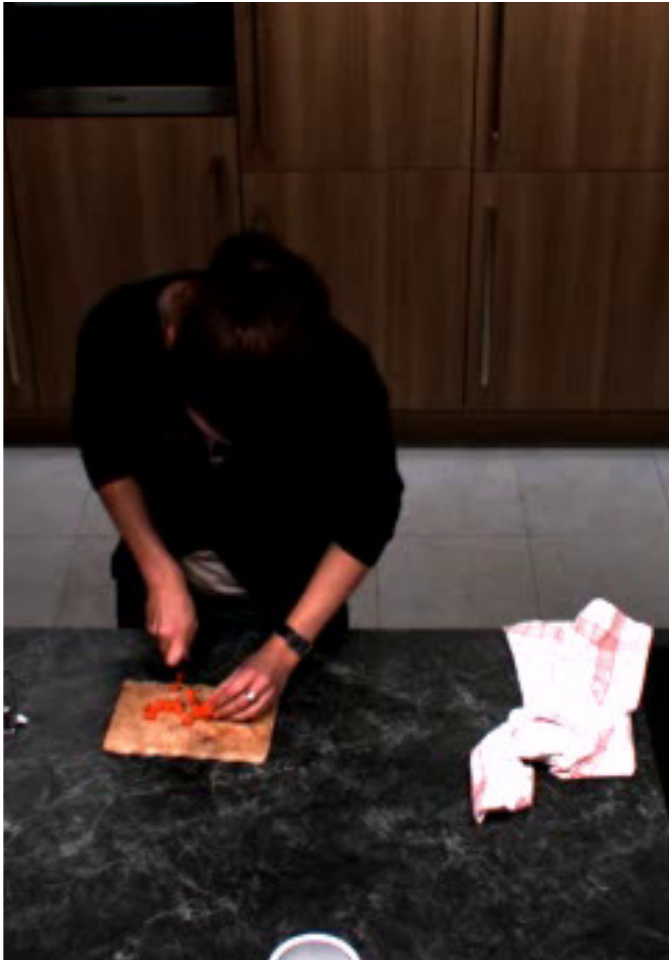}\\
      \end{tabular}\\      
      \begin{tabular}{l|l}
        RNN-sent:
        &
        \begin{tabular}{l}
          \textit{The person took out a cutting board from the drawer.}\\
          {\color{Red}\textit{The person got a knife and a cutting
              board from the drawer.}}\\
          \textit{The person cut the ends off the cutting board.}\\
        \end{tabular}\\
        \hline
        h-RNN:
        &
        \begin{tabular}{l}
          \textit{The person took out a cutting board.}\\
          {\color{Green}\textit{The person got a knife from the drawer.}}\\
          \textit{The person cut the cucumber on the cutting
            board.}\\
        \end{tabular}\\
      \end{tabular}
    \end{tabular}
  }
  \vspace*{1ex}
  \caption{Examples of generated paragraphs.
    {\color{Red}Red} indicates incorrect sentences produced by
    RNN-sent and {\color{Green}green} shows the ones generated by our
    h-RNN in the corresponding time intervals.
    In the first example, our hierarchical model successfully
    captures the high likelihood of the event \textit{walk to the sink}
    after the event \textit{open the refrigerator}. 
    In the second example, RNN-sent generates the event \textit{take
      the cutting board} twice due to the fact that the sentences in the
    paragraph are produced independently.
    In contrast, our hierarchical model avoids this mistake.
  }
  \label{fig:examples}
  \vspace{-2ex}
\end{figure}

The results on TACoS-MultiLevel are shown in Table~\ref{tab:tacos}.
Our approach outperforms the state-of-the-art methods, including the
very recently proposed one (\ie, LRCN) with an improvement of
$\frac{0.305-0.292}{0.292}=4.45\%$ in the BLEU@4 score.
Given that our strategy of extracting object regions is relatively
simple compared to the sophisticated hand detector
\cite{Donahue2014,Rohrbach2014}, we expect to have even better
performance if our object localization is improved.
Our method is also superior to all the baseline methods.
Although RNN-cat models temporal dependency among sentences by
sentence-level concatenation, it performs worse than our hierarchical
architecture.
Again, it shows that both the video features and the hierarchical
structure are crucial in our task.
Figure~\ref{fig:examples} illustrates some example paragraphs
generated by our approach on TACoS-MultiLevel.


To further demonstrate that our method h-RNN generates better
sentences than RNN-cat in general, we perform human evaluation to
compare these two methods on TACoS-MultiLevel.
Specifically, we discard $1,166$ test video intervals, each of which
has exactly the same sentence generated by RNN-cat and h-RNN.
This results in a total number of $4,314-1,166=3,148$ video intervals
for human evaluation.
We then put the video intervals and the generated sentences on Amazon
Mechanical Turk (AMT).
Each video interval is paired with one sentence generated by RNN-cat
and the other by h-RNN, side by side.
For each video interval, we ask one turker to select the sentence that
better describes the video content.
The turker also has a third choice if he believes that both sentences
are equally good or bad.
In the end, we obtained $773$ selections for h-RNN and $472$
selections for RNN-cat, with a gap of $301$ selections.
Thus h-RNN has at least $\frac{301}{472+3069}=8.50\%$ improvement over
RNN-cat.
\begin{table}[h]
  \centering
  \vspace*{-2ex}
  \begin{tabular}{c|c|c|c}
    h-RNN & RNN-cat & Equally good or bad & Total\\
    \hline
    773 & 472 & 3069 & 4314\\
  \end{tabular}
\end{table}

\vspace*{-4ex}
\subsection{Discussions and Limitations}
Although our approach is able to produce paragraphs for video and has
achieved encouraging results, it is subject to several limitations.
First, our object detection routine has difficulty handling very small
objects.
Most of our failure cases on TACoS-MultiLevel produce incorrect object
names in the sentences, \eg, confusing small objects that have similar
shapes or appearances (\textit{cucumber vs.\@ carrot}, \textit{mango
  vs.\@ orange}, \textit{kiwi vs.\@ avocado}, \etc).
See Figure~\ref{fig:cherry-pick} for a concrete example:
\textit{sliced the orange} should really be \textit{sliced the mango}.
Accurately detecting small objects (sometimes with occlusion) in
complex video scenarios still remains an open problem.
Second, the sentential information flows unidirectionally through the
paragraph recurrent layer, from the beginning of the paragraph to the
end, but not also in the reverse way.
Misleading information will be potentially passed down when the first
several sentences in a paragraph are generated incorrectly.
Using bidirectional RNN~\cite{Schuster1997,Wen2015} for sentence
generation is still an open problem.
Lastly, our approach suffers from a known problem as in most other
image/video captioning methods, namely, there is discrepancy between
the objective function used by training and the one used by
generation.
The training process predicts the next word given the previous words
from groundtruth, while the generation process conditions the
prediction on the ones previously generated by itself.
This problem is amplified in our hierarchical framework where the
paragraph generator conditions on groundtruth sentences during
training but on generated ones during generation.
A potential cure for this would be adding Scheduled
Sampling~\cite{Bengio2015} to the training process, where one randomly
selects between the true previous words and the words generated by the
model.
Another solution might be to directly optimize the metric (\eg, BLEU)
used at test time~\cite{Ranzato2015}.

\vspace{-1ex}
\section{Conclusion}
We have proposed a hierarchical-RNN framework for video paragraph
captioning.
The framework models inter-sentence dependency to generate a sequence
of sentences given video data.
The experiments show that our approach is able to generate a paragraph
for a long video and achieves the state-of-the-art results on two
large-scale datasets.

\vspace{-1ex}
\section*{Acknowledgments}
\vspace{-1ex}
The primary author would like to thank Baidu Research for providing
the summer internship.

\end{document}